\documentclass[sigconf ,prologue,table]{acmart}
\copyrightyear{2023}
\acmYear{2023}
\setcopyright{acmlicensed}
\acmConference[KDD '23] {Proceedings of the 29th ACM SIGKDD Conference on Knowledge Discovery and Data Mining}{August 6--10, 2023}{Long Beach, CA, USA.}
\acmBooktitle{Proceedings of the 29th ACM SIGKDD Conference on Knowledge Discovery and Data Mining (KDD '23), August 6--10, 2023, Long Beach, CA, USA}
\acmPrice{15.00}
\acmISBN{979-8-4007-0103-0/23/08}
\acmDOI{10.1145/3580305.3599353}

\usepackage{amsfonts}
\usepackage{bbm}
\usepackage[linesnumbered,ruled]{algorithm2e}
\usepackage{optidef}
\usepackage{amsmath}
\usepackage{mathtools}
\usepackage{hyperref}
\usepackage{stmaryrd}
\usepackage{enumitem}
\usepackage{balance}

\usepackage[linesnumbered,ruled]{algorithm2e}
\SetAlCapNameFnt{\scriptsize}
\SetAlCapFnt{\scriptsize}

\SetKwInput{KwInput}{Input}                
\SetKwInput{KwOutput}{Output}  
\SetKwRepeat{Repeat}{repeat}{until}
\SetKwRepeat{RepeatFor}{repeat}{for}

\title{\textsc{Fire}: An Optimization Approach for Fast Interpretable Rule Extraction}

\author{Brian Liu}
\affiliation{%
  \institution{Massachusetts Institute of Technology }
   \city{Cambridge}
   \state{Massachusetts}
  \country{USA}}
\email{briliu@mit.edu}

\author{Rahul Mazumder}
\affiliation{%
  \institution{Massachusetts Institute of Technology }
   \city{Cambridge}
   \state{Massachusetts}
  \country{USA}}
\email{rahulmaz@mit.edu}
\renewcommand{\shortauthors}{Brian Liu \& Rahul Mazumder} 
\begin{document}

\begin{abstract}

We present \textsc{Fire}, Fast Interpretable Rule Extraction, an optimization-based framework to extract a small but useful collection of decision rules from tree ensembles. \textsc{Fire} selects sparse representative subsets of rules from tree ensembles, that are easy for a practitioner to examine. To further enhance the interpretability of the extracted model, \textsc{Fire} encourages fusing rules during selection, so that many of the selected decision rules share common antecedents. 
The optimization framework utilizes a fusion regularization penalty to accomplish this, along with a non-convex sparsity-inducing penalty to aggressively select rules. Optimization problems in \textsc{Fire} pose a challenge to off-the-shelf solvers due to problem scale and the non-convexity of the penalties. To address this, making use of problem-structure, we develop a specialized solver based on block coordinate descent principles; our solver performs up to 40$\times$ faster than existing solvers. We show in our experiments that \textsc{Fire} outperforms state-of-the-art rule ensemble algorithms at building sparse rule sets, and can deliver more interpretable models compared to existing methods.

\end{abstract}



\maketitle
\section{Introduction}

Tree ensembles are popular for their versatility and excellent off-the-shelf performance. While powerful, these models can grow to massive sizes and become difficult to interpret. To improve model parsimony and interpretability, decision rules can be extracted from trained tree ensembles. Each leaf node in a decision tree represents a decision rule; the path of internal nodes from root to leaf in the tree forms a conjunction of if-then antecedents that assigns a prediction to a partition of the dataset. Extracting a sparse (or parsimonious) subset of decision rules (leaf nodes) from a tree ensemble can produce a compact and transparent model that performs well in terms of prediction accuracy \cite{friedman2008predictive}. 

\begin{figure}[h]
    \centering
    \includegraphics[width = 0.45\textwidth]{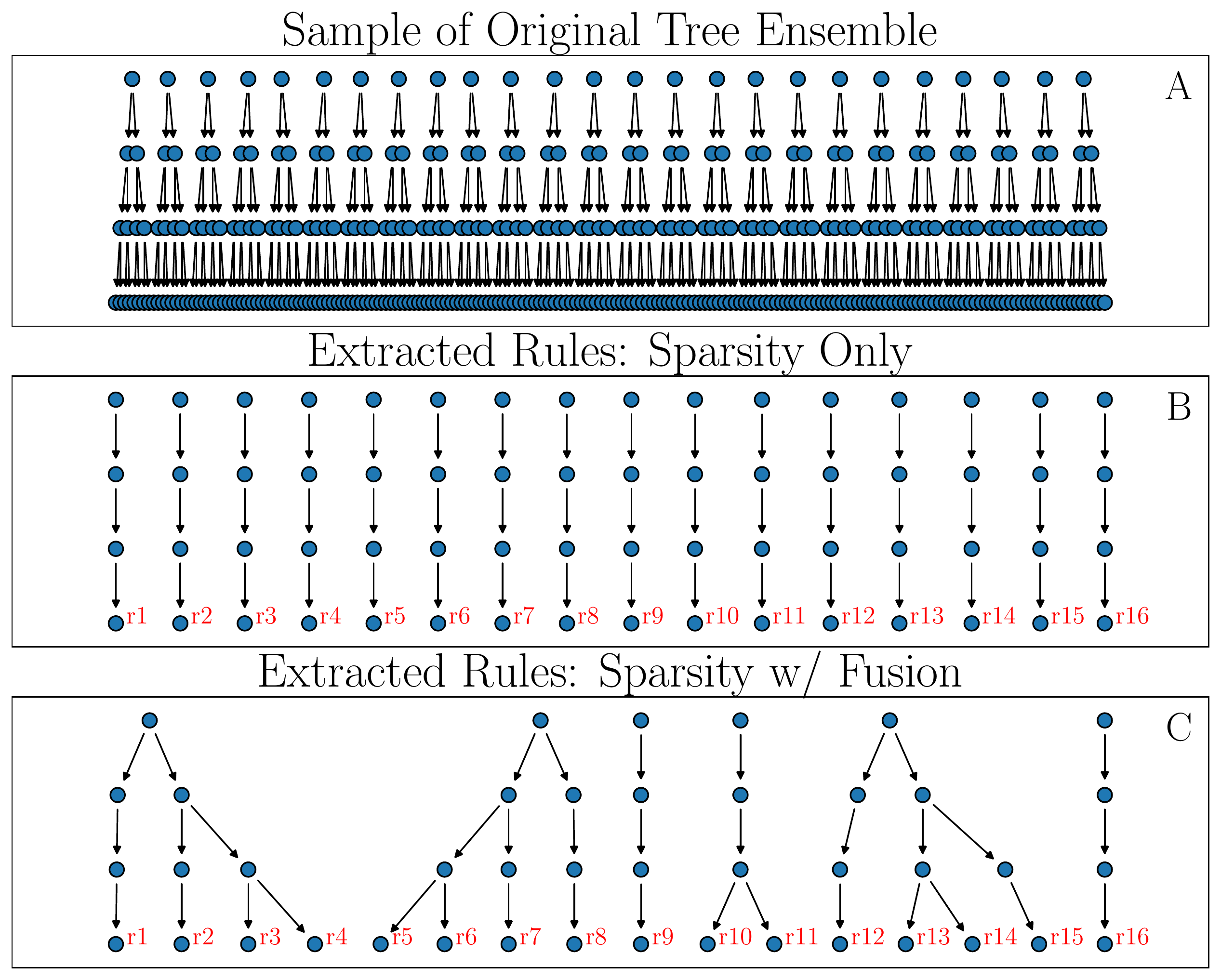}
    \caption{Fusion improves parsimony by reducing internal nodes. In both panels, 16 decision rules are selected but the sparsity with fusion panel contains \textbf{44\%} fewer internal nodes.}
    \label{intro_example.fig}
\end{figure}

In this paper, we present the Fast Interpretable Rule Extraction (\textsc{Fire}) framework, an optimization-based framework to extract an interpretable collection of rules from tree ensembles. The goal of \textsc{Fire} is to select a small subset of decision rules that is representative of the larger collection of rules found in a tree ensemble. In addition to sparsity, \textsc{Fire} allows for the flexibility to encourage \emph{fusion} in the extracted rules. In other words, the framework can encourage the selection of multiple rules that are {\emph{close together}} from within the same decision tree, so that the selected rules (leaf nodes) share common antecedents (internal nodes). As we discuss later, encouraging fusion appears to improve the parsimony and interpretability of the extracted rule ensemble. To better convey our intuition,  Figure \ref{intro_example.fig} presents an illustration. From the original tree ensemble (panel A), we extract 16 decision rules by encouraging only sparsity (panel B), and by encouraging fusion with sparsity (panel C)\footnote{This example is based off an application of our framework.}. The 16 decision rules selected in the sparsity-only panel each come from a different decision tree while the decision rules selected in the fusion with sparsity panel come from only 6 trees. As a result, the rule set extracted by encouraging both fusion and sparsity contains substantially fewer internal nodes, since leaf nodes from the same decision tree share internal nodes. This translates to fewer if-then antecedents for a practitioner to examine in the rule ensemble, suggesting improved interpretability.

\textsc{Fire} is based on an optimization formulation that assigns a weight to each decision rule in a tree ensemble and extracts a sparse subset of rules by minimizing regularized loss function. This allows a practitioner to evaluate the trade-off between model compactness and performance by varying the regularization penalty, and to select an appropriately-sized model. \textsc{Fire} uses a non-convex sparsity-inducing penalty popularly used in highdimensional linear models to aggressively select rules and fused LASSO penalty \cite{tibshirani2005sparsity} to encourage rule fusion. The fused LASSO is a classical tool used in the context of approximating a signal via a piecewise constant approximation using an $\ell_1$-based penalty---we present a novel exploration of this tool in the context of rule ensemble extraction.








Optimization problems in \textsc{Fire} pose a challenge to existing solvers due to problem size and the non-convexity of the penalties. 
On that account, we develop a novel optimization algorithm to efficiently obtain high-quality solutions to these optimization problems. Our algorithms leverage problem structure and block coordinate descent combined with greedy selection heuristics to improve computational efficiency. By exploiting the blocking structure of problems in \textsc{Fire}, our specialized solver scales and allows for computation that appear to be well beyond the capabilities of off-the-shelf solvers. In addition, our algorithms support warm start continuation across tuning parameters, which allows a practitioner to use \textsc{Fire} to rapidly extract rule sets of varying sizes.

With our specialized solver, \textsc{Fire} is computationally fast and easy to tune. We show in our experiments that \textsc{Fire} extracts sparse rule sets that outperform state-of-the-art competing algorithms, by up to a $24\%$ decrease in test error. We also demonstrate through a real-world example that \textsc{Fire} extracts decision rules that are easier to interpret compared to the rules selected by existing methods.

Our paper is organized as follows. We first overview rule extraction from tree ensembles. We then introduce our model framework and discuss the effects of our new penalties. We next present our specialized optimization algorithm along with timing experiments against off-the-shelf solvers. Finally, we present our experimental results and our interpretability case study. An open-source implementation of \textsc{Fire} along with a supplement containing derivations and experimental details can be found in this project repository\footnote{\href{https://github.com/brianliu12437/FIREKDD2023}{https://github.com/brianliu12437/FIREKDD2023}}.

\subsection{Main Contributions}

\begin{itemize}[noitemsep,topsep=0pt,parsep=0pt,partopsep=0pt, leftmargin=*]
    \item We introduce the \textsc{Fire} framework for rule extraction. \textsc{Fire} selects sparse representative subsets of decision rules from tree ensembles and can encourage fusion so that the selected rules share common antecedents.
    \item \textsc{Fire} is based on a regularized loss minimization framework. Our regularizer comprises of a non-convex sparsity-inducing penalty to aggressively select rules, and a fused LASSO penalty to encourage fusion. Our work is the first to explore this family of penalty functions originating in high-dimensional statistics in the context of rule extraction.
    \item We show how encouraging fusion (in addition to vanilla sparsity) when extracting rules improves the interpretability and compression of the selected model.
    \item Optimization problems in \textsc{Fire} are challenging due to problem scale and the non-convex penalties, so we develop a specialized solver for our framework. Our algorithm computes solutions up to \textbf{40$\times$} faster than off-the-shelf solvers on medium-sized problems (10000s of data points and decision variables) and can scale to larger problems.
    \item We show in our experiments that \textsc{Fire} extracts sparse rule sets that outperform rule ensembles built by state-of-the-art algorithms, with up to a \textbf{24\%} decrease in test error. In addition, \textsc{Fire} performs significantly better than RuleFit \cite{friedman2008predictive}, a classical optimization-based rule extraction algorithm, with up to a \textbf{46\%} decrease in test error when extracting sparse models.
\end{itemize}

\section{Preliminaries \& Related Work}
In this section provide a cursory overview of decision trees, rules, and tree ensembles, and survey existing work on rule extraction.

\subsection{Decision Trees and Decision Rules}

Given feature matrix $X \in \mathbb{R}^{N \times P}$ and target $y \in \mathbb{R}^{N}$, decision tree $\Gamma(X)$ maps $\mathbb{R}^{N \times P} \rightarrow \mathbb{R}^N$. A decision tree of maximum depth $d$ partitions the training data into at most $2^{d}$ non-overlapping partitions. Each partition, or leaf node, is defined by a sequence of at most $d$ splits and data points in a partition are assigned the mean (regression) or majority class (classification) for predictions. Each split is an if-then rule that thresholds a single feature; splits partition the data based on whether the feature value of a data point falls above or below that threshold. 

Decision rules are conjunctions of if-then antecedents that partition a dataset and assign a prediction to each partition \cite{friedman2008predictive}. Decision trees can be viewed as a collection of decision rules, where each leaf node in the tree is a rule. The decision path to each leaf node is a conjunction of if-then antecedents and data points partitioned by these antecedents are assigned a prediction equal to the mean or majority class of the node.
\begin{figure}[h!]
  \begin{minipage}[c]{0.25\textwidth}
    \includegraphics[width=\textwidth]{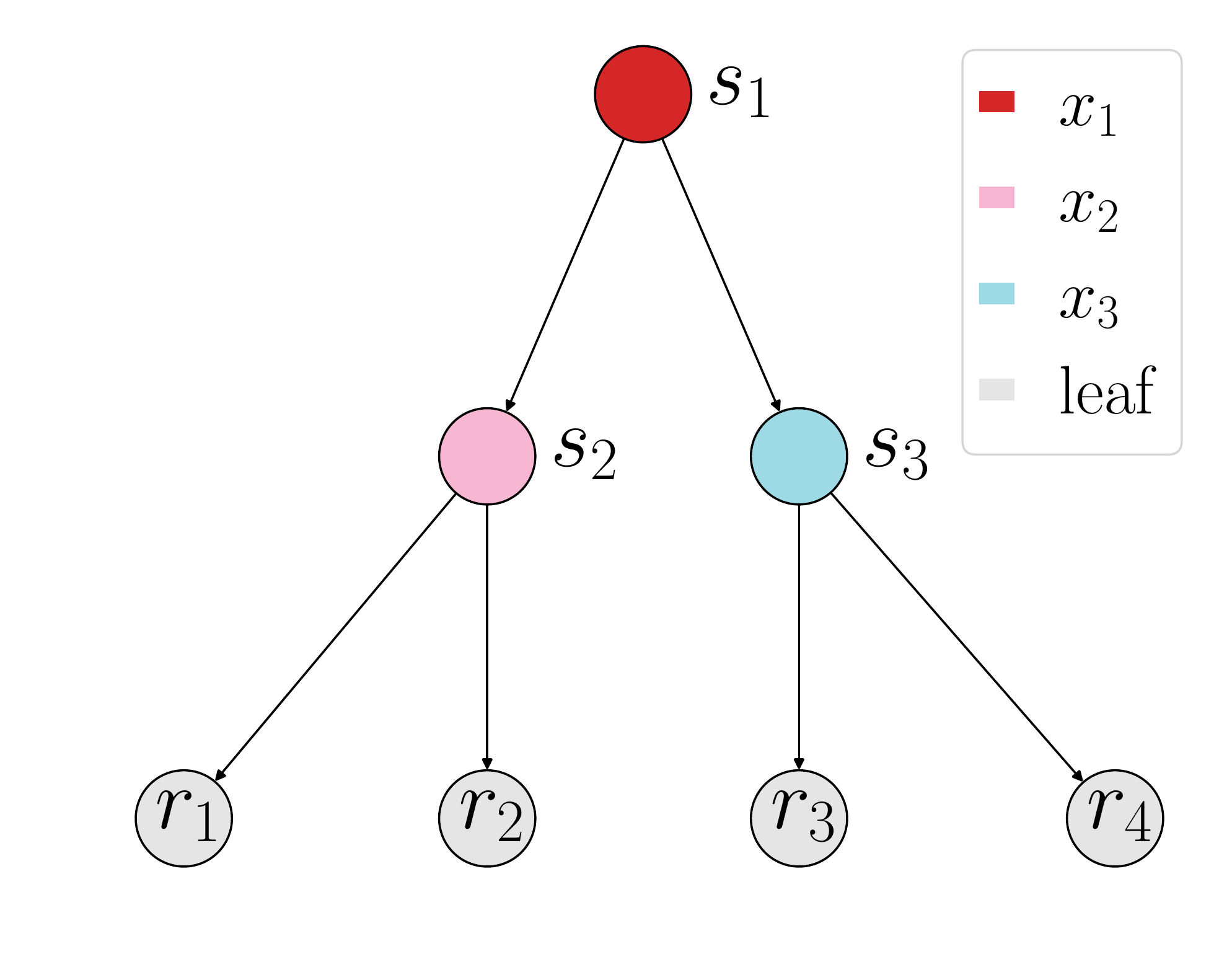}
  \end{minipage}\hfill
  \begin{minipage}[c]{0.2\textwidth}
    \caption{This depth 2 decision tree yields 4 decision rules. The antecedents of each rule are obtained by traversing the tree from root to leaf.
     } \label{tree_to_rules.fig}
  \end{minipage}
\end{figure}
For example, consider the decision tree shown in figure \ref{tree_to_rules.fig}. Let $s_j$ denote the threshold for the split on feature $x_j$ and for each split let $x_j \leq s_j$ denote the left path and let $x_j > s_j$ denote the right path. The rule obtained from leaf node $r_3$ can be represented as:
    $r_3(x) =\mathbbm{1}(x_1 > s_1 ) \cdot \mathbbm{1}( x_3 \leq s_3  ) \cdot v_3$, where $\mathbbm{1}(x)$ is the indicator function and $v_3$ is the prediction value of leaf node $r_3$. More generally, the rule obtained from a leaf node whose decision path traverses $S$ splits can be expressed by:
    $r(x) = \prod_{j = 1} ^ S \mathbbm{1}(x \in \sigma_j) \cdot v$, where $\sigma_j$ is the set of data points partitioned along the decision path by split $j$ and $v$ is the value of the node.

\subsection{Tree Ensembles}

Tree ensembles consist of a collection of $T$ decision trees, $\{\Gamma_t(X): t \in [T]\}$. This collection of trees can be obtained via \emph{bagging} \cite{breiman1996bagging}, where trees are trained in parallel on bootstrapped samples of the data, or through \emph{boosting} \cite{friedman2001greedy}, where dampened trees are added sequentially and trained on the residuals of the prior ensemble. 

Rule ensembles (rule sets) are collections of decision rules with weights assigned to each rule. The prediction of a rule ensemble is obtained by taking the weighted linear combination of the rules. For example, we can obtain a rule ensemble from the decision tree in figure \ref{tree_to_rules.fig} by assigning each rule $r_j(x)$ weight $w_j$. The prediction of the rule ensemble can be expressed as: $\sum_{j=1}^{4} w_j r_j(x).$

Tree ensembles result in large collections of decision rules which can be extracted into sparse rule ensembles. We use the following notation to discuss extracting rule ensembles from trees. Consider a decision tree $\Gamma_t$, fit on data matrix $X \in \mathbb{R}^{N \times P}$, with $R_t$ leaf nodes. Each leaf node has prediction value $v_j$ for $j \in [R_t]$. Recall that if data point $x_i$ reaches leaf node $r_j$, the data point is assigned prediction value $v_j$. 
We define a mapping matrix $M_t \in  \mathbb{R}^{N \times R_t}$ whose $(i,j)$-th entry is given by:
\begin{equation}
    (M_t)^{ij} = \begin{cases}
    v_j & \text{if data point $x_i$ reaches leaf node $r_j$} \\
    0 & \text{otherwise}.
    \end{cases}
\end{equation}
Mapping matrix $M_t$ maps data points to predictions. The matrix is sparse, with density $\frac{1}{R_t}$, since each data point is routed to a single leaf in the decision tree. Let weight vector $w_t \in \mathbb{R}^{R_t}$ represent the weights assigned to each leaf node; $M_t$ and $w_t$ define the rule ensemble obtained from $\Gamma_t$. The prediction of this rule ensemble is given by $M_t w_t$. 

For an ensemble of $T$ trees, define $M_t$ for each tree $t \in [T]$. Let $R = \sum_{t=1}^T R_t$ denote the total number of rules (nodes) in the ensemble and denote the mapping matrix $M \in \mathbb{R}^{N \times R}$ as $M = [M_1, M_2, \ldots M_T].$
Matrix $M$ is also sparse with density $\frac{T}{R}$. Given weight vector, $w \in \mathbb{R}^R$, the prediction of this rule ensemble is $Mw$. To extract rules, we fit weight vector $w$; setting an entry of $w$ to zero prunes the corresponding rule from the ensemble.

\subsection{Related Work}

Extracting decision rules from tree ensembles was first introduced in 2005 by RuleFit \cite{friedman2008predictive}. Following the notation introduced above, RuleFit selects a subset of rules by minimizing the $\ell_1$-regularized optimization problem (aka LASSO): 
\begin{mini}|s|
{w}{(1/2)\left\|y-M w\right\|_{2}^{2} + \lambda \left\|w \right\|_1,}{\label{rulefit_problem}}{}
\end{mini}
which penalizes the $\ell_1$ norm of the weights $w$. RuleFit uses LASSO solvers to compute a solution $w$ to Problem~\eqref{rulefit_problem}.

Subsequently, various algorithms to post-process or generate rule ensembles have been proposed. Node Harvest \citep{meinshausen2010node} uses the non-negative garrote and quadratic programming to select rule ensembles from tree ensembles. More recently, SIRUS \citep{benard2021sirus} uses stabilized random forests to build and aggregate rule sets, and GLRM \citep{wei2019generalized} uses column generation to create rules from scratch. To the best of our knowledge, \textsc{Fire} is the first framework that incorporates improved sparse selection and rule groupings within a holistic optimization framework. We show in our experiments that \textsc{Fire} outperforms SIRUS and GLRM at selecting sparse human-readable rule sets.

\begin{figure}[h]
    \centering
    \includegraphics[width = 0.45\textwidth]{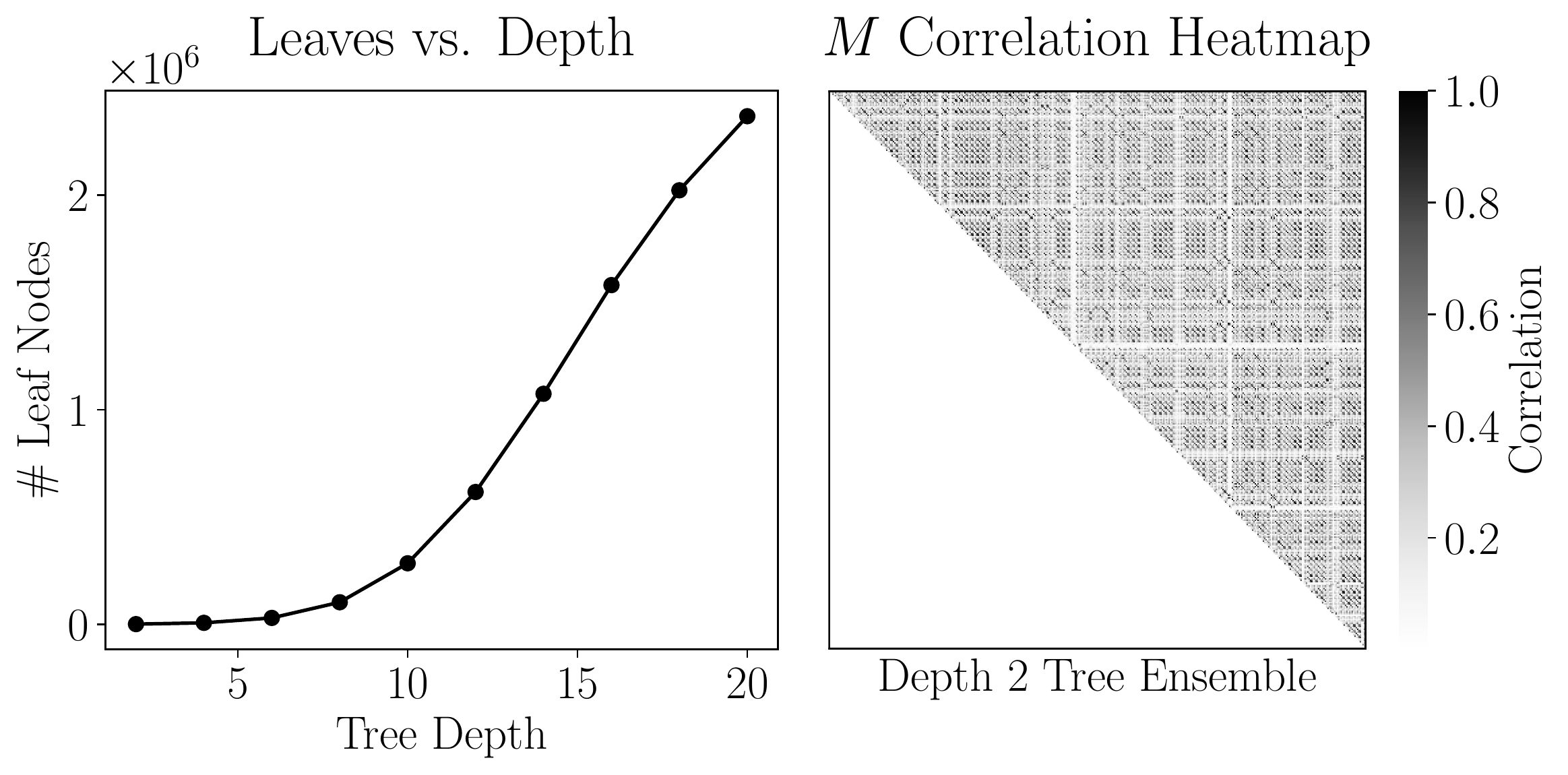}
    \caption{Bagging ensemble of 500 trees fit on the Elevators dataset \cite{OpenML2013}. The number of decision rules in the ensemble scales exponentially with tree depth. Shallower ensembles contain highly correlated rules.} \label{challenges_example.fig}
\end{figure}

Existing optimization-based rule extraction algorithms, such as RuleFit, face several challenges due to the structure of tree ensembles. The number of variables in the optimization problem (i.e. the number of leaves in the tree ensemble) increases exponentially with tree depth, as shown in the left plot in figure \ref{challenges_example.fig}. RuleFit uses cyclic coordinate descent to solve the LASSO, which becomes expensive when the number of coordinates is large. As a result, RuleFit is restricted for use on shallow tree ensembles. We show in \S\ref{optimization_algo.section} that our specialized optimization algorithm for \textsc{Fire} is robust to the depth of the ensemble and scales substantially better than the LASSO solvers used by RuleFit.

An important difference between \textsc{Fire} and RuleFit stems from a simple yet critical observation: shallow tree ensembles contain many correlated decision rules.
The right plot in figure \ref{challenges_example.fig} shows the pairwise correlations between the columns of M on a depth $2$ ensemble of 500 trees; many pairs of columns have correlation scores close to 1. This further complicates rule extraction, since LASSO performs poorly at sparse selection on highly correlated features \cite{hebiri2012correlations, sun2020correlated,hazimeh2022l0learn}. Earlier work in high-dimensional statistics proposes the use of non-convex penalties, which performs better at sparse selection in the presence of correlation \cite{zhang2010nearly,mazumder2011sparsenet}.

\section{Proposed Modeling Framework}
\label{framework.section}
In this section, we present our model framework and discuss in detail the effects of our non-convex sparsity-inducing penalty and fusion penalty on decision rule extraction.

Consider a tree ensemble with $T$ decision trees $\{\Gamma_t(X): t \in [T]\}$. Each decision tree has $R_t$ leaf nodes and mapping matrix $M_t \in \mathbb{R}^{N \times R_t}$. The ensemble has $R = \sum_{t=1}^T R_t$ leaf nodes and mapping matrix $M \in \mathbb{R}^{N \times R}$. \textsc{Fire} selects a sparse subset of rules by learning weights $w$ by solving:
\begin{mini}|s|
{w}{f(w) +  h(w, \lambda_s) + g(w, \lambda_f), }{\label{main_prob}}{}
\end{mini}
where $f(w) = (1/2)\left\|y-M w\right\|_{2}^{2}$ is quadratic loss that measures data-fidelity, $h$ is the sparsity penalty with regularization parameter $\lambda_s$, and $g$ is the fusion penalty with parameter $\lambda_f$.

\subsection{Sparsity-Inducing Penalty}

We discuss possible choices for sparsity-inducing penalty $h$. One baseline choice is the LASSO from RuleFit where, $h(w, \lambda_s) = \lambda_s \sum_{j=1}^R |w_j|$. However, this $\ell_1$-penalty encourages heavy shrinkage and bias in $w$, which makes the LASSO a poor choice for sparse variable selection in the presence of correlated variables \cite{hebiri2012correlations,hazimeh2022l0learn}. 
Since we intend to perform sparse selection from a collection of correlated rules, we use a non-convex penalty which incurs less bias than LASSO.

Many unbiased and nearly unbiased penalty functions exist, such as the $\ell_0$-penalty \cite{hazimeh2022l0learn}, the smoothly clipped absolute deviation (SCAD) penalty \cite{fan2001variable}, and the minimax concave plus (MCP) penalty \cite{zhang2010nearly}.
\textsc{Fire} uses the MCP penalty since it is continuous and sub-differentiable---properties that will come in handy when we develop our optimization solver. We set $h$ as:
\begin{equation}
    h(w, \lambda_s) = \sum_{j=1}^R P_{\gamma}(w_j, \lambda_s),
\end{equation}
where $P_{\gamma}(w_j, \lambda_s)$ is the MCP penalty function defined by:
\begin{equation}
    P_{\gamma}(w_j,\lambda_s) = \begin{cases} 
\lambda_s |w_j|  - \frac{{w_j}^2}{2 \gamma} &\text{if } |w_j| \leq \lambda_s \gamma \\
 \frac{1}{2}\gamma \lambda_s^2 & \text{if } |w_j| > \lambda_s \gamma, \end{cases}
\end{equation}
and $\gamma > 1$ is a hyperparameter that (loosely speaking) controls the concavity of the penalty function. As $\gamma \sim \infty$, the penalty behaves like the LASSO, and when $\gamma \sim 1^+$ it operates like the $\ell_0$-penalty.



\subsection{Fusion Penalty}
\label{fusion.section}
In addition to sparse selection, we also present a framework 
to encourage rule fusion. To this end, we use a fused LASSO penalty \citep{tibshirani2005sparsity} to encourage contiguity in the leaf nodes (rules) selected or pruned from within each tree. The fused LASSO penalizes the sum of absolute differences between the coefficients and is commonly used for piecewise constant signal approximation \cite{hoefling2010path}. Here, we explore how this classical penalty function can be used to improve rule extraction.

Let $w_t \in \mathbb{R}^{R_t} $ represent the sub-vector of weights in $w$ that correspond to tree $\Gamma_t$ and let $(w_t)_j$ denote the $j$-th entry of $w_t$. We set $g$ as:
\begin{equation}
    g(w, \lambda_f) = \lambda_f \sum_{t=1}^T \left\| D_t w_t \right\|_1
\end{equation}
where $D_t \in \{\text{-} 1, 0 ,1 \}^{( R_t - 1) \times R_t}$ is the tree fusion matrix with $(D_t)_{ij} = -1$ for all $i = j$ and $(D_t)_{ij} = 1$ for all $i = j - 1$. We have that:
\begin{equation}
    \left\|D_t w_t\right\|_1 = \sum_{j = 2}^{R_t} | (w_t)_j - (w_t)_{j-1}|.
    \label{fusion_penalty.eq}
\end{equation}
This penalizes the absolute value of the differences of the fitted weights in each tree. As a result, in problem~\eqref{main_prob}, for larger values of $\lambda_f$, the nodes assigned zero weights and pruned (and the nodes assigned non-zero weights and selected) are grouped together in each tree. In what follows we provide some intuition into why this can be appealing to a practitioner. 

\subsubsection{Compressing Trees}
\begin{figure}[h!]
    \centering
    \includegraphics[width = 0.45\textwidth ]{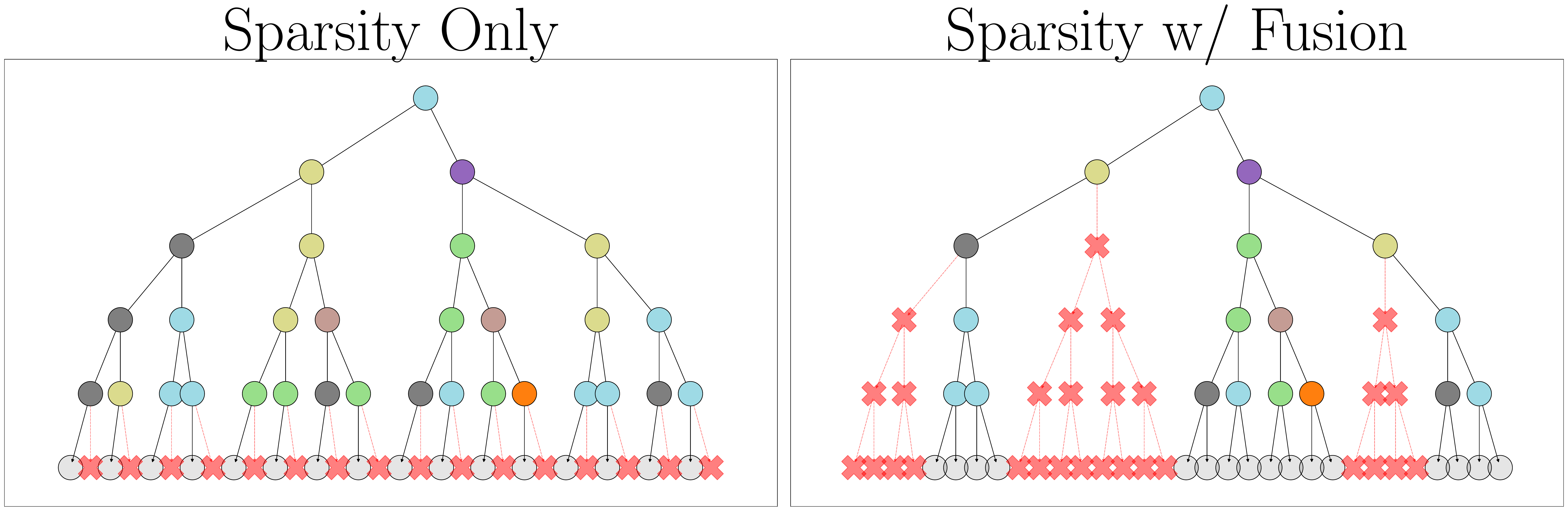}
    \caption{Grouping pruned leaves increases the number of internal nodes removed from a decision tree.}
    \label{fusion_compression.fig}
\end{figure}

By grouping pruned leaf nodes together, we increase the number of internal nodes removed from a tree, since an internal node whose children are pruned is removed as well. Consider this example in figure \ref{fusion_compression.fig}.

In both plots, we prune 16 out of the 32 leaf nodes from a depth 5 decision tree fit on the California Housing Prices dataset \citep{OpenML2013}. In the left plot the pruned leaves are noncontiguous so \textbf{0} internal nodes are removed; the pruned tree contains \textbf{47} total (leaf + internal) nodes. In contrast, the pruned leaves are grouped in the right plot. Consequently, \textbf{13} additional internal nodes are removed and the pruned tree contains \textbf{34} total nodes. Both trees incur the same rule sparsity penalty of 16 leaves, but the right tree contains 28\% fewer total nodes. The fusion penalty $g$ encourages grouping in the leaf nodes pruned from each tree which further compresses tree ensembles compared to only regularizing for sparsity in the leaves.

\begin{figure}[h!]
    \centering
    \includegraphics[width = 0.45\textwidth ]{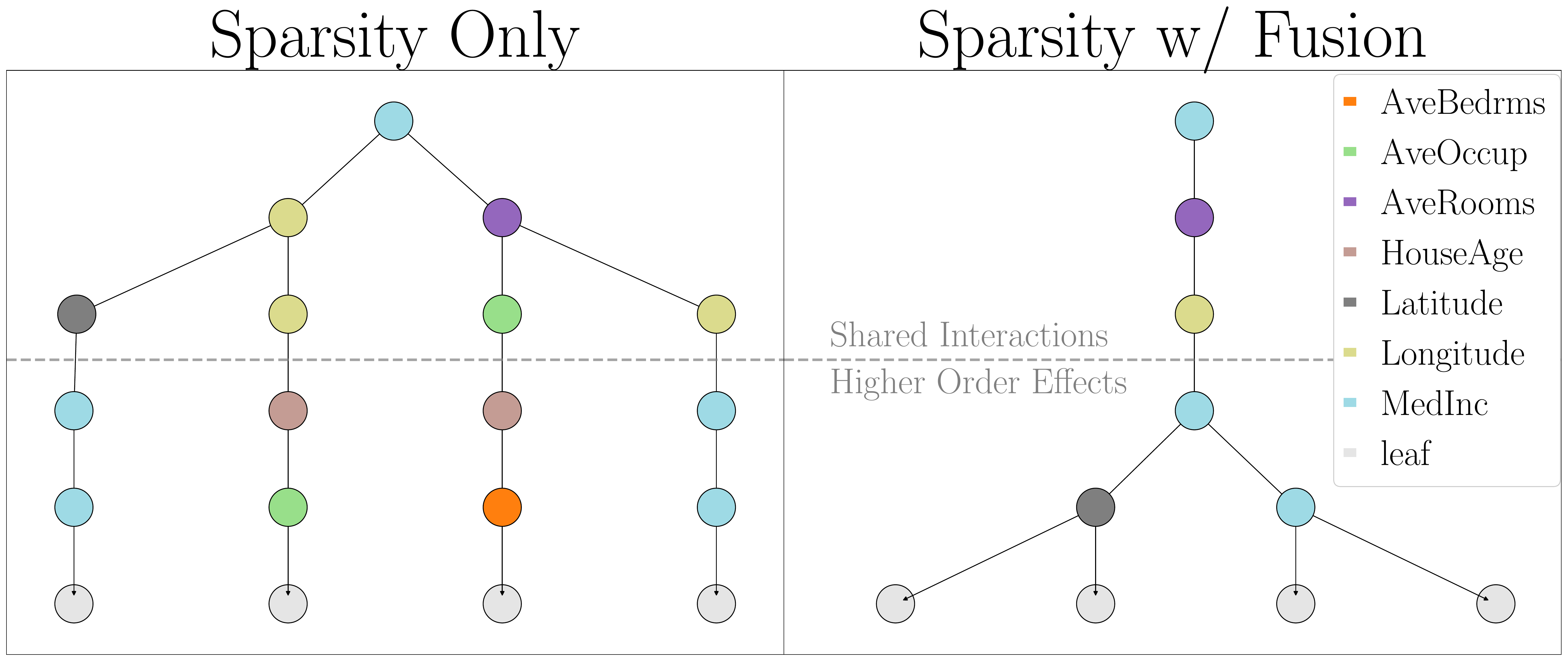}
    \caption{Extracting 4 rules from a decision tree fit on the California Housing Price dataset. In the right plot, the rules are grouped together and are more interpretable.}
    \label{fusion_groupings.fig}
\end{figure}

\subsubsection{Grouping Rules}
Grouping the rules selected from each tree improves model interpretability since grouped rules share antecedents. In figure \ref{fusion_groupings.fig}, we select 4 rules from the decision tree shown in figure \ref{fusion_compression.fig}. The left plot selects rules that are spread out and the right plot selects rules that are grouped.

Consider the task of interpreting all interaction effects in the rule ensemble up to depth 3. The 4 rules in the right ensemble share the antecedents: $\mathbbm{1}(\text{MedInc} > s_1)$, $\mathbbm{1}(\text{AveRooms} > s_2)$, and $\mathbbm{1}( \text{Longitude}  > s_3)$. A user would need to analyze \textbf{3} antecedents to interpret the interactions. For the left ensemble, a user would need to analyze \textbf{7} antecedents to interpret all depth 3 interactions.

Fusion regularizer $g$ \eqref{fusion_penalty.eq} introduces a more natural way of penalizing the complexity of the selected rules. Consider selecting two leaf nodes; the first leaf node shares a parent node with a leaf that has already been selected and the second leaf node is on a branch where no leaves have been selected. Selecting the first leaf adds no internal nodes (antecedents) to the rule ensemble while selecting the second leaf can add up to $d$ new antecedents. Sparsity regularizer $h$ penalizes both choices equally but the fusion regularizer incurs an additional penalty for the second choice.

\subsection{Hyperparameters}
\label{param_selection.section}

We discuss how to select good values for hyperparameters in \textsc{Fire}. We denote $\lambda_s$ as the sparsity hyperparameter, $\gamma$ as the concavitiy hyperparameter, and $\lambda_f$ as the fusion hyperparameter.

\subsubsection{Sparsity}
Sparsity hyperparameter  $\lambda_s$ generally controls the number of rules extracted from the tree ensemble, i.e., the number of nonzero entries in $w$. We can use warm start continuation to efficiently compute the entire regularization path of $w$ across $\lambda_s$ with the other hyperparameters held fixed. We start with a value of $\lambda_s$ sufficiently large such that $w^* = \textbf{0}$ and decrement $\lambda_s$, using the previous solution as a warm start to Problem \ref{main_prob}, until the full model is reached \citep{friedman2007pathwise}. This procedure allows a practitioner to quickly evaluate rule ensembles of different sizes. Given any fixed configuration of $\gamma$ and $\lambda_f$, it is easy to select $\lambda_s$; we compute the regularization path for the hyperparameter and find the value of $\lambda_s$ that minimizes validation loss.

\subsubsection{Concavity}

Concavity hyperparameter $\gamma$ controls the trade-off between shrinkage and selection in the MCP penalty. When $\gamma \xrightarrow[]{} 1^{+}$, the MCP penalty aggressively performs nearly unbiased selection. When $\gamma \xrightarrow[]{} \infty$, the MCP penalty encourages regularization through shrinkage.

\begin{figure}[h!]
    \centering
    \includegraphics[width = 0.45\textwidth ]{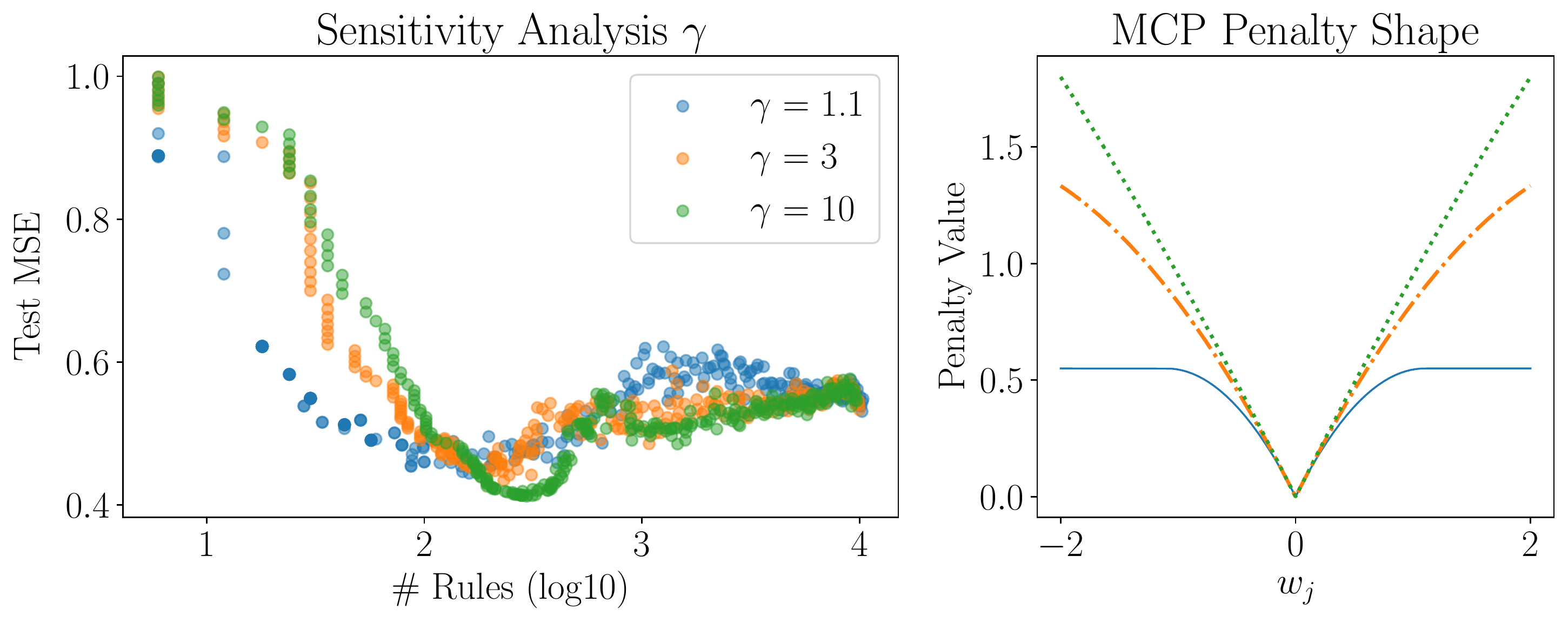}
    \caption{Effect of $\gamma$ on the MCP penalty. Varying $\gamma$ controls the trade-off between shrinkage and selection.}
    \label{sensitivity_analysis.fig}
\end{figure}

The left plot in figure \ref{sensitivity_analysis.fig} demonstrates this effect on extracting rules from a tree ensemble fit on the US Communities and Crime dataset \cite{OpenML2013}. We conduct a sensitivity analysis on the MCP penalty by varying $\gamma \in \{1.1, 3, 10\}$ and computing the regularization path of $\lambda_s$ for each value of $\gamma$. The horizontal axis shows the number of rules extracted 
and the vertical axis shows the test performance of the selected model. The right plot in figure \ref{sensitivity_analysis.fig} shows the corresponding shape of the MCP penalty function.

When $\gamma \xrightarrow[]{} 1^{+}$, the MCP penalty performs better at selecting very sparse subsets of rules, due to the reduced bias of the more concave penalty \cite{zhang2010nearly}. However, this aggressive selection can possibly result in overfitting in low-signal regimes. Increasing $\gamma$ increases the shrinkage imparted by the MCP penalty, which regularizes the model and reduces overfitting. As $\gamma$ increases, the model is less likely to overfit in the low regularization regime (RHS of figure \ref{sensitivity_analysis.fig}) but performs worse at sparse selection (LHS of figure \ref{sensitivity_analysis.fig}).

Selecting the best value for $\gamma$ depends on the use case for \textsc{Fire}. When using \textsc{Fire} to select very sparse rule ensembles, we recommend  setting $\gamma$ to be close to $1^{+}$. Otherwise, 
we consider a small number of possible
$\gamma$ values and for each value, we compute the regularization path for $\lambda_s$. We use validation-tuning to select an appropriate $(\lambda_s, \gamma)$ pair.
Our framework's ability to use warm-start continuation to compute regularization paths for $\lambda_s$ makes this 2-dimensional tuning computationally efficient.


\subsubsection{Fusion}

Fusion hyperparameter $\lambda_f$ influences the interpretability of the extracted rule ensemble. Increasing $\lambda_f$ encourages more fused rules which are easier to interpret. The best value of $\lambda_f$ is use-case specific, and we observe empirically that values of $\lambda_f \in [0.5 \lambda_s , 2 \lambda_s]$ work well. We show in our case study in \S\ref{interpretability_case.section} the effect of $\lambda_f$ on the interpretability of the selected ensemble. Increasing $\lambda_f$ also adds additional regularization, which may be useful for preventing overfitting on noisy datasets.

\section{ Optimization Algorithm}
\label{optimization_algo.section}
We present our specialized optimization algorithm to efficiently obtain high-quality solutions to problem \eqref{main_prob}. Note that the smooth loss function and non-smooth regularizers in problem \eqref{main_prob} are separable across blocks $w_t$'s, where each block represents a tree in the ensemble. Motivated by the success of block coordinate descent (BCD) algorithms for large-scale sparse regression \citep{friedman2010regularization, hazimeh2022l0learn}, we apply the method to problem \eqref{main_prob}. As we discuss below, our proposed algorithm has notable differences: we make use of a block structure to perform updates---taking advantage of a structure that naturally arises from the tree ensemble. Also, a direct application of cyclic coordinate descent approaches can be quite expensive, so we use a greedy selection rule motivated by the success of greedy coordinate descent for LASSO problems \cite{karimireddy2019efficient,fang2020greed}. This results in important computational savings, as our experiments in \S\ref{computation_time_experiment} show.


\subsection{Block Proximal Updates}\label{blockupdate.section}
We make use of a natural blocking structure that arises in our tree ensemble. Specifically, each block $t$ corresponds to a tree in the ensemble with mapping matrix $M_t$ and associated weights $w_t \in \mathbb{R}^{R_t}$. For a fixed block $t$, let $\delta$ denote the other blocks. The goal of a block update is to update weights $w_t$ while holding everything else constant. The optimization criterion for each block update is:
\begin{mini}|s|
{w_t}{f(w_t) +  h(w_t, \lambda_s) + g(w_t, \lambda_f)}{\label{blockupdate_problem}}{}
\end{mini}
where  $f(w_t) = \frac{1}{2} \left\| \hat{y}_{\delta} - M_t w_t \right\|_2^2$ and $\hat{y}_{\delta} = y - M_{\delta} w_{\delta}$. This composite criterion has smooth loss function $f$ and non-smooth regularizers $h$ and $g$, so we apply (block) proximal gradient updates~\citep{beck2009fast,nesterov2018lectures}.

The function $w_{t} \mapsto f(w_t)$ has Lipschitz continuous gradient and satisfies $\left\| \nabla f(u) - \nabla f(v) \right\| \leq L_t \left\| u - v \right\|$,
for all $u$ and $v$, where $L_t$ is the largest eigenvalue of $M_t^\intercal M_t$. At point $w_t^k$ each proximal update minimizes the quadratic approximation of objective 
\eqref{blockupdate_problem} and can be expressed as:
\begin{argmini}|s|{\theta}{(L_t/{2}) \| \theta - \hat{\theta} \|_2^2 +  h(\theta, \lambda_s) + g(\theta, \lambda_f),}{\label{blockupdate_proximal_obj}}{w_t^{k+1} = }
\end{argmini}
where $\hat{\theta} = w_t^k - \frac{1}{L_t} \nabla f(w_t^k)$. Our choice of step size $\frac{1}{L_t}$ ensures that the proximal updates lead to a sequence of decreasing objective values \citep{beck2009fast}. We show that univariate problem \eqref{blockupdate_proximal_obj} has closed-form minimizers $\theta^*$ for all choices of sparsity penalty $h$ when $g = 0$, and can be rapidly solved using dynamic programming when fusion penalty $g$ is introduced.

\subsubsection{Sparsity Only}
Consider the case where $\gamma_f = 0$ and $g(\theta,\gamma_f) = 0$. For $h(\theta, \lambda_s) = \lambda_s \left\|\theta \right\|_1$, the closed-form minimzer to problem \eqref{blockupdate_proximal_obj} is equal to $\theta^* = S_{(\lambda_s/L_t)}(\hat{\theta})$, where $S_{\lambda}$ is the soft-thresholding operator given elementwise by: $S_{\lambda}(\hat{\theta}_j) = \text{sign}(\hat{\theta}_j)(|\hat{\theta}_j| - \lambda|)_+$.
This expression is obtained through computing subgradient optimality conditions for problem \eqref{blockupdate_proximal_obj} \citep{beck2009fast}. We can repeat this procedure with $h(\theta, \lambda_s)$ as the MCP sparsity penalty. The closed-form minimizer to problem \eqref{blockupdate_proximal_obj} is given elementwise by,
\begin{equation}
\label{mcp_threshold_operator}
    \theta_j^* = \begin{cases}
                \frac{\gamma}{\gamma - 1} S_{\frac{\lambda_s}{L_t}}(\hat{\theta}_j) & \text{if } |\hat{\theta}_j| \leq  \frac{\lambda_s \gamma}{L_t} \\
                \hat{\theta}_j & \text{if } |\hat{\theta}_j| >  \frac{\lambda_s \gamma}{L_t}.
                \end{cases}
\end{equation}
Denote expression \eqref{mcp_threshold_operator} as the MCP thresholding operator.
As $\gamma \xrightarrow[]{} \infty$ the operator behaves like soft-thresholding and as $\gamma \xrightarrow[]{} 1^{+}$ the operator behaves like hard-thresholding. Derivations for both minimizers are presented in the supplement (suppl. C.1 \& C.2).

\subsubsection{Sparsity with Fusion} Now consider the case where the fusion penalty $g$ is nonzero. We start with sparsity penalty set to zero: $\lambda_s = 0$ and $h(\theta,\lambda_s) = 0$. Problem \eqref{blockupdate_proximal_obj} can be re-expressed as:
\begin{argmini}|s|{\theta}{(1/2) \| \theta - \hat{\theta} \|_2^2  + (\lambda_f/L_t) \left\|D_t \theta \right\|_1, }{\label{fused_proximal_obj}}{}
\end{argmini}
which is equivalent to the 1-dimensional fused lasso signal approximation problem (FSLA), with fusion penalty parameter $\frac{\lambda_f}{L_t}$. This FSLA problem can be solved efficiently using the dynamic programming algorithm proposed by \cite{johnson2013dynamic}, in linear worst-case time complexity with respect to the number of training observations.

Given the solution to problem \eqref{fused_proximal_obj}, $\theta^*(0,\lambda_f)$,
we can find the solution to problem \eqref{blockupdate_proximal_obj}, $\theta^*(\lambda_s,\lambda_f)$, for any $\lambda_s > 0$ by applying the soft-thresholding operator to  $\theta^*(0,\lambda_f)$ if $h$ is the $\ell_1$-penalty, or by applying the MCP thresholding operator if $h$ is the MCP penalty. We derive these procedures in the supplement (suppl. C.3 \& C.4).

For improved computational performance, we conduct 5-10 proximal gradient iterations for each block update by solving \eqref{blockupdate_proximal_obj}. This problem either has a closed-form minimizer or can be solved in O($N$) time complexity, so blocks can be efficiently updated in constant or linear time.  In the following section, we present a method to prioritize the selection of blocks.

\subsection{Block Selection}
We first discuss unguided block selection methods. Cyclic block selection cycles through blocks $\{1 \ldots T\}$ and updates them one at a time until convergence, while randomized block selection updates a random block at each iteration. BCD algorithms that use unguided block selection are typically slow; guided greedy block selection can greatly reduce computation time \citep{nutini2015coordinate}. We present a novel greedy block selection heuristic for problem \eqref{main_prob}.

Greedy selection uses heuristics to find the best block or coordinate to update at each iteration. For example, the Gauss Southwell steepest direction (GS-s) rule picks the steepest coordinate as the best coordinate to update. For smooth functions, this corresponds to the coordinate with the largest gradient magnitude. For composite functions, the steepest direction is computed with respect to the subgradients of the regularizers \citep{karimireddy2019efficient}. For our composite objective in problem \eqref{main_prob} define the  direction vector $d \in \mathbb{R}^R$ elementwise by:
 \begin{mini}|s|
{s \in \partial h_j + \partial g_j}{| \nabla_j f(w) + s |.}{\label{steepest_direction_vector}}{d_j =}
\end{mini} 
The GS-s rule selects the entry of $d$ with the largest magnitude as the best coordinate to update. To find the best block to update, we modify the GS-s rule following Nutini et al. (2017)\cite{nutini2017let} and select the block whose direction vector has the largest magnitude. Let $[T]$ represent the set of all blocks and $d_t \in \mathbb{R}^{R_t}$ represent the elements of $d$ associated with block $t$. Select the best block to update $t^*$ via:
 \begin{argmaxi}|s|
{t \in [T]}{\left\| d_{t} \right\|.}{\label{steepest_block_selection}}{t^* = }
 \end{argmaxi}
 Problems \eqref{steepest_direction_vector} and \eqref{steepest_block_selection} form our greedy block selection heuristic. Our heuristic is only useful if problem \eqref{steepest_direction_vector} can be efficiently solved. Karimireddy et al. (2019) \cite{karimireddy2019efficient} derives a closed-form minimizer for this problem  when $h$ is the $\ell_1$-penalty and $g=0$. Our algorithm is novel in that we derive closed-form minimizers to find $d$ when fusion penalty $g$ is introduced, and when $h$ is the MCP penalty. This requires computing the subgradients of a modified fused LASSO problem \citep{tibshirani2011solution} and the MCP penalty function; the derivation is lengthy and is presented for all penalties in the supplement (suppl. D).
 
\subsubsection{Discussion}
\begin{figure}[h]
    \centering
    \includegraphics[width = 0.45\textwidth ]{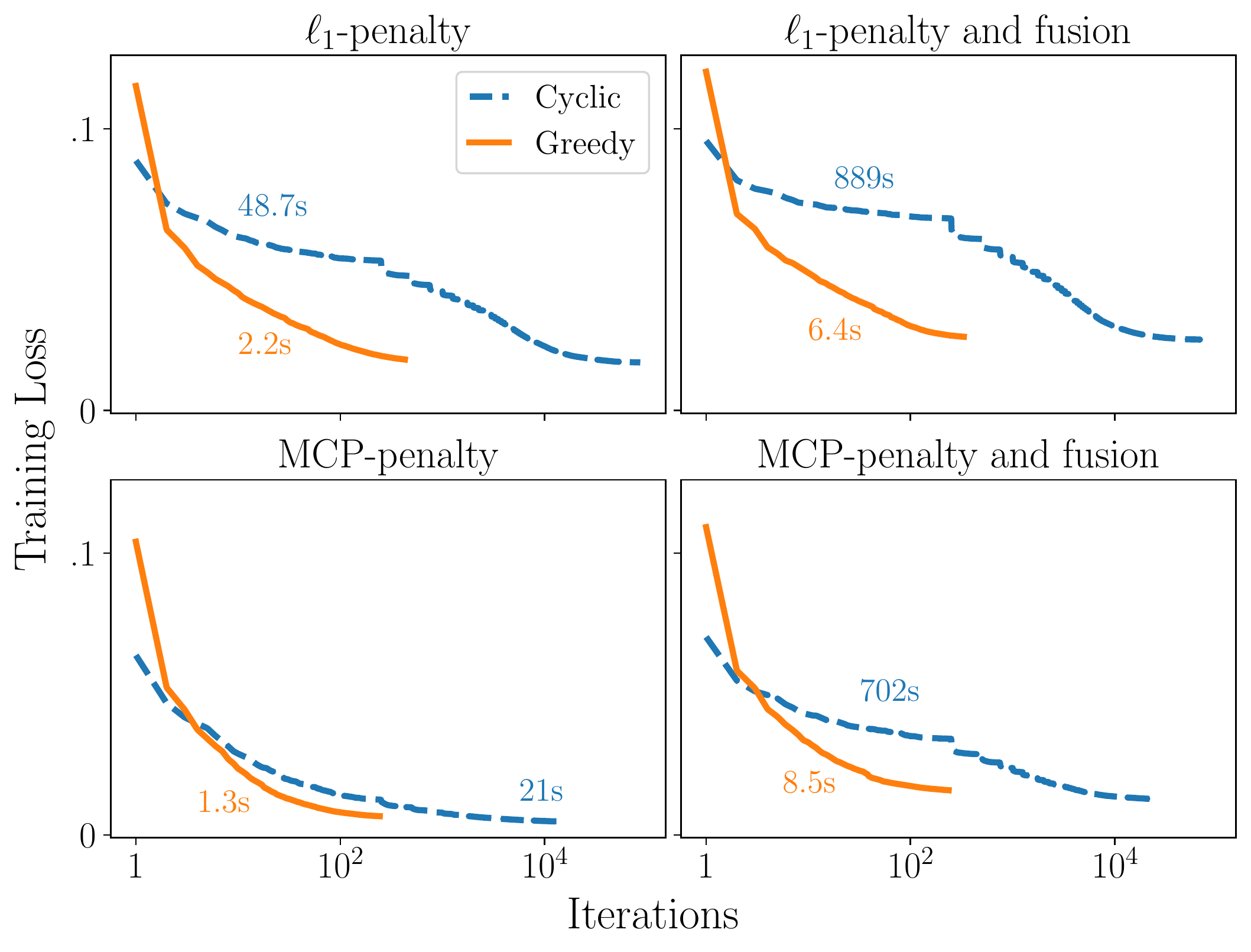}
    \caption{Training loss v. \# iterations for greedy v. cyclic BCD, with computation time in seconds. Greedy block selection takes 2 orders of magnitude (\textbf{100x}) fewer iterations. Horizontal axis is log scale.}
    \label{greedy_v_cyclic.fig}
\end{figure}

Our greedy block selection heuristic drastically reduces the number of BCD iterations as we demonstrate below. Fit a tree ensemble of 250 trees (blocks) with 7500 leaves (rules) on the Stock Price Prediction dataset \citep{OpenML2013}, which contains 1000 rows.
Select a sparse rule ensemble from~\eqref{main_prob} for various choices of $h$ and $g$, with $\lambda_s = 1$, $\lambda_f = 0.5$, and $\gamma = 1.1$. Compare the progress of our algorithm using cyclic block selection versus greedy block selection. From figure \ref{greedy_v_cyclic.fig} we observe that greedy BCD requires 2 orders of magnitude fewer iterations compared to cyclic BCD. Greedy BCD iterations are costlier than cyclic BCD iterations since finding the steepest direction vector at each iteration requires computing the full gradient. However, greedy BCD drastically reduces computation time. Here, greedy BCD takes \textbf{8.5} seconds for~\eqref{main_prob} with the MCP and fusion penalty, while cyclic BCD takes \textbf{702} seconds. The timing results for the other configurations are shown in figure \ref{greedy_v_cyclic.fig}.

\subsection{Putting Together the Pieces}
\begin{algorithm}
\scriptsize 
\caption{Greedy Block Coordinate Descent (GBCD) Solver  }
\label{gbcdalgo}
\DontPrintSemicolon
  
  \KwInput{y, \ $M$, \ $\lambda_s$, \ $\lambda_f$, \ $\gamma$, \ $[T]$}
  
  \textbf{Initialize} $w = \textbf{0}$ 
 
  \Repeat{objective no longer improves}{
  
  Find steepest direction vector $d$: \textcolor{blue}{Problem \eqref{steepest_direction_vector}}
  
  Select block $t$: \textcolor{blue}{Problem \eqref{steepest_block_selection}}

  \RepeatFor{5 to 10 iterations}{
  Proximal block update $w_{t}$: \textcolor{blue}{Problem \eqref{blockupdate_proximal_obj}}
  } }

\textbf{optional} sweep through all blocks to check convergence. \label{optional_convergence_check}
  
  \KwOutput{w}
\end{algorithm}

Algorithm \ref{gbcdalgo} presents our greedy block coordinate descent (GBCD) algorithm. Our algorithm is efficient; the block selection problems have closed-form minimizers and the block update problems either have closed-form minimizers or can be solved in linear $0(N)$ worst-case time complexities. We include step \ref{optional_convergence_check} as an optional step where we conduct cyclic block coordinate descent sweeps to ensure that our algorithm converges. In practice, we observe that usually only a single pass over the blocks is needed to verify convergence.

\subsection{Computation Time Experiment}
\label{computation_time_experiment}
Here we compare the computation time of GBCD against existing off-the-shelf optimization solvers. Since the MCP and fusion penalties are novel to our framework, we use GBCD to solve problem \eqref{main_prob} with $h$ set as the $\ell_1$-penalty and $g = 0$. 
This optimization problem is the same as the one in RuleFit, so we can directly compare our GBCD algorithm against existing LASSO solvers.

\subsubsection{Medium-Sized Problems}
We build a random forest of 250 trees grown to depths 3, 5, and 7 and use GBCD to sparsify the ensemble with $\lambda_s = 0.1$. Under this configuration, GBCD typically selects 20\% of the rules. This represents a realistic use case for RFOP; users compute the regularization path up to some sparsity level and select the best model. With $\lambda_s = 0.1$, we show the computation time required for a single solve near the middle of the path.

We compare GBCD against the Python implementation of RuleFit \citep{molnar_2016}, which uses the LASSO coordinate descent solver in Scikit-learn \citep{scikit-learn}. Also, we compute the full regularization path for $\lambda_s \in [0.01,1000]$ using our GBCD algorithm with warm start continuation and the LASSO coordinate path function in Scikit-learn and compare the timing results. Finally, we compare GBCD for a single solve against Node Harvest implemented using CVXPY \citep{diamond2016cvxpy}. Node Harvest solves a different optimization problem than GBCD but we include this algorithm to compare GBCD against other optimization-based rule extraction methods. We conduct this timing experiment on a personal laptop with a 2.80 GHz Intel processor.

\begin{table}[h]
\scalebox{0.6} {
\begin{tabular}{|cccc|cccccc}
\cline{1-4} \cline{6-9}
\multicolumn{4}{|c|}{\textbf{Rule Extraction Single Solve}}                                                                                                                                                                                                                                                                                                                             & \multicolumn{1}{c|}{} & \multicolumn{4}{c|}{\textbf{Rule Extraction Full Path}}                                                                                                                                                                                                                                                                                &  \\ \cline{1-4} \cline{6-9}
\multicolumn{4}{|c|}{\cellcolor[HTML]{C0C0C0}{\color[HTML]{333333} \textbf{GBCD Single Solve}}}                                                                                                                                                                                                                                                                                                  & \multicolumn{1}{c|}{} & \multicolumn{4}{c|}{\cellcolor[HTML]{C0C0C0}{\color[HTML]{333333} \textbf{GBCD Regularization Path}}}                                                                                                                                                                                                                                           &  \\ \cline{1-4} \cline{6-9}
\multicolumn{1}{|c|}{\textbf{\begin{tabular}[c]{@{}c@{}}Rows/\\ Vars\end{tabular}}} & \multicolumn{1}{c|}{\textbf{\begin{tabular}[c]{@{}c@{}}2000\\  (3)\end{tabular}}}                        & \multicolumn{1}{c|}{\textbf{\begin{tabular}[c]{@{}c@{}}8000 \\ (5)\end{tabular}}}                        & \textbf{\begin{tabular}[c]{@{}c@{}}25000 \\ (7)\end{tabular}}                        & \multicolumn{1}{c|}{} & \multicolumn{1}{c|}{\textbf{\begin{tabular}[c]{@{}c@{}}Rows/\\ Vars\end{tabular}}} & \multicolumn{1}{c|}{\textbf{\begin{tabular}[c]{@{}c@{}}2000\\  (3)\end{tabular}}} & \multicolumn{1}{c|}{\textbf{\begin{tabular}[c]{@{}c@{}}8000 \\ (5)\end{tabular}}} & \multicolumn{1}{c|}{\textbf{\begin{tabular}[c]{@{}c@{}}25000 \\ (7)\end{tabular}}} &  \\ \cline{1-4} \cline{6-9}
\multicolumn{1}{|c|}{\textbf{1316}}                                                 & \multicolumn{1}{c|}{\cellcolor[HTML]{A1E9A1}{\color[HTML]{333333} 21.6 (0.6)}}                           & \multicolumn{1}{c|}{\cellcolor[HTML]{A1E9A1}{\color[HTML]{333333} 16.5 (1.4)}}                           & \cellcolor[HTML]{A1E9A1}{\color[HTML]{333333} 16.3 (0.4)}                            & \multicolumn{1}{c|}{} & \multicolumn{1}{c|}{\textbf{1316}}                                                 & \multicolumn{1}{c|}{\cellcolor[HTML]{A1E9A1}{\color[HTML]{333333} 90.8 (2.0)}}    & \multicolumn{1}{c|}{\cellcolor[HTML]{A1E9A1}{\color[HTML]{333333} 83.8 (5.5)}}    & \multicolumn{1}{c|}{\cellcolor[HTML]{A1E9A1}{\color[HTML]{333333} 90.4 (7.4)}}     &  \\ \cline{1-4} \cline{6-9}
\multicolumn{1}{|c|}{\textbf{4338}}                                                 & \multicolumn{1}{c|}{\cellcolor[HTML]{A1E9A1}{\color[HTML]{333333} 63.9 (2.2)}}                           & \multicolumn{1}{c|}{\cellcolor[HTML]{A1E9A1}{\color[HTML]{333333} 137.7 (2.2)}}                          & \cellcolor[HTML]{A1E9A1}{\color[HTML]{333333} 146.4 (0.9)}                           & \multicolumn{1}{c|}{} & \multicolumn{1}{c|}{\textbf{4338}}                                                 & \multicolumn{1}{c|}{\cellcolor[HTML]{A1E9A1}{\color[HTML]{333333} 196.0 (2.0)}}   & \multicolumn{1}{c|}{\cellcolor[HTML]{A1E9A1}{\color[HTML]{333333} 465.2 (10.9)}}  & \multicolumn{1}{c|}{\cellcolor[HTML]{A1E9A1}{\color[HTML]{333333} 653.25 (21.8)}}  &  \\ \cline{1-4} \cline{6-9}
\multicolumn{1}{|c|}{\textbf{10955}}                                                & \multicolumn{1}{c|}{\cellcolor[HTML]{A1E9A1}{\color[HTML]{333333} 106.4 (3.9)}}                          & \multicolumn{1}{c|}{\cellcolor[HTML]{A1E9A1}{\color[HTML]{333333} 176.8 (0.6)}}                          & \cellcolor[HTML]{A1E9A1}{\color[HTML]{333333} 419.6 (1.6)}                           & \multicolumn{1}{c|}{} & \multicolumn{1}{c|}{\textbf{10955}}                                                & \multicolumn{1}{c|}{\cellcolor[HTML]{A1E9A1}{\color[HTML]{333333} 350.0 (5.4)}}   & \multicolumn{1}{c|}{\cellcolor[HTML]{A1E9A1}{\color[HTML]{333333} 732.8 (14.8)}}  & \multicolumn{1}{c|}{\cellcolor[HTML]{A1E9A1}{\color[HTML]{333333} 1712.3 (37.2)}}  &  \\ \cline{1-4} \cline{6-9}
\multicolumn{4}{|c|}{\cellcolor[HTML]{C0C0C0}\textbf{Scikit-learn RuleFit}}                                                                                                                                                                                                                                                                                                                      & \multicolumn{1}{c|}{} & \multicolumn{4}{c|}{\cellcolor[HTML]{C0C0C0}\textbf{Scikit-learn RuleFit Regularization Path}}                                                                                                                                                                                                                                                  &  \\ \cline{1-4} \cline{6-9}
\multicolumn{1}{|c|}{\textbf{\begin{tabular}[c]{@{}c@{}}Rows/\\ Vars\end{tabular}}} & \multicolumn{1}{c|}{{\color[HTML]{333333} \textbf{\begin{tabular}[c]{@{}c@{}}2000\\  (3)\end{tabular}}}} & \multicolumn{1}{c|}{{\color[HTML]{333333} \textbf{\begin{tabular}[c]{@{}c@{}}8000 \\ (5)\end{tabular}}}} & {\color[HTML]{333333} \textbf{\begin{tabular}[c]{@{}c@{}}25000 \\ (7)\end{tabular}}} & \multicolumn{1}{c|}{} & \multicolumn{1}{c|}{\textbf{\begin{tabular}[c]{@{}c@{}}Rows/\\ Vars\end{tabular}}} & \multicolumn{1}{c|}{\textbf{\begin{tabular}[c]{@{}c@{}}2000\\  (3)\end{tabular}}} & \multicolumn{1}{c|}{\textbf{\begin{tabular}[c]{@{}c@{}}8000 \\ (5)\end{tabular}}} & \multicolumn{1}{c|}{\textbf{\begin{tabular}[c]{@{}c@{}}25000 \\ (7)\end{tabular}}} &  \\ \cline{1-4} \cline{6-9}
\multicolumn{1}{|c|}{\textbf{1316}}                                                 & \multicolumn{1}{c|}{{\color[HTML]{333333} 221.9 (12.4)}}                                                 & \multicolumn{1}{c|}{{\color[HTML]{333333} 492.3 (19.8)}}                                                 & \cellcolor[HTML]{FFFFFF}{\color[HTML]{333333} 628.3 (32.7)}                          & \multicolumn{1}{c|}{} & \multicolumn{1}{c|}{\textbf{1316}}                                                 & \multicolumn{1}{c|}{{\color[HTML]{333333} 1250.9 (17.6)}}                         & \multicolumn{1}{c|}{{\color[HTML]{333333} 1732.2 (19.2)}}                         & \multicolumn{1}{c|}{\cellcolor[HTML]{FFCCC9}{\color[HTML]{333333} $>$ 1800}}       &  \\ \cline{1-4} \cline{6-9}
\multicolumn{1}{|c|}{\textbf{4338}}                                                 & \multicolumn{1}{c|}{\cellcolor[HTML]{FFFFFF}{\color[HTML]{333333} 998.4 (130.5)}}                        & \multicolumn{1}{c|}{\cellcolor[HTML]{FFCCC9}{\color[HTML]{333333} $>$ 1800}}                             & \cellcolor[HTML]{FFCCC9}{\color[HTML]{333333} $>$ 1800}                              & \multicolumn{1}{c|}{} & \multicolumn{1}{c|}{\textbf{4338}}                                                 & \multicolumn{1}{c|}{\cellcolor[HTML]{FFCCC9}{\color[HTML]{333333} $>$ 1800}}      & \multicolumn{1}{c|}{\cellcolor[HTML]{FFCCC9}{\color[HTML]{333333} $>$ 1800}}      & \multicolumn{1}{c|}{\cellcolor[HTML]{FFCCC9}{\color[HTML]{333333} $>$ 1800}}       &  \\ \cline{1-4} \cline{6-9}
\multicolumn{1}{|c|}{\textbf{10955}}                                                & \multicolumn{1}{c|}{\cellcolor[HTML]{FFCCC9}{\color[HTML]{333333} $>$ 1800}}                             & \multicolumn{1}{c|}{\cellcolor[HTML]{FFCCC9}{\color[HTML]{333333} $>$ 1800}}                             & \cellcolor[HTML]{FFCCC9}{\color[HTML]{333333} $>$ 1800}                              & \multicolumn{1}{c|}{} & \multicolumn{1}{c|}{\textbf{10955}}                                                & \multicolumn{1}{c|}{\cellcolor[HTML]{FFCCC9}{\color[HTML]{333333} $>$ 1800}}      & \multicolumn{1}{c|}{\cellcolor[HTML]{FFCCC9}{\color[HTML]{333333} $>$ 1800}}      & \multicolumn{1}{c|}{\cellcolor[HTML]{FFCCC9}{\color[HTML]{333333} $>$ 1800}}       &  \\ \cline{1-4} \cline{6-9}
\multicolumn{4}{|c|}{\cellcolor[HTML]{C0C0C0}\textbf{Node Harvest (CVXPY ECOS)}}                                                                                                                                                                                                                                                                                                                 &                       & \multicolumn{4}{c}{\cellcolor[HTML]{FFFFFF}\textbf{}}                                                                                                                                                                                                                                                                                           &  \\ \cline{1-4}
\multicolumn{1}{|c|}{\textbf{\begin{tabular}[c]{@{}c@{}}Rows/\\ Vars\end{tabular}}} & \multicolumn{1}{c|}{{\color[HTML]{333333} \textbf{\begin{tabular}[c]{@{}c@{}}2000\\  (3)\end{tabular}}}} & \multicolumn{1}{c|}{{\color[HTML]{333333} \textbf{\begin{tabular}[c]{@{}c@{}}8000 \\ (5)\end{tabular}}}} & {\color[HTML]{333333} \textbf{\begin{tabular}[c]{@{}c@{}}25000 \\ (7)\end{tabular}}} &                       & \multicolumn{4}{c}{\cellcolor[HTML]{FFFFFF}\textbf{}}                                                                                                                                                                                                                                                                                           &  \\ \cline{1-4}
\multicolumn{1}{|c|}{\textbf{1316}}                                                 & \multicolumn{1}{c|}{{\color[HTML]{333333} 114.2 (5.6)}}                                                  & \multicolumn{1}{c|}{{\color[HTML]{333333} 109.1 (3.2)}}                                                  & {\color[HTML]{333333} 297.3 (6.0)}                                                   &                       & {\color[HTML]{333333} \textbf{}}                                                   & \textbf{}                                                                         & \textbf{}                                                                         & \textbf{}                                                                          &  \\ \cline{1-4}
\multicolumn{1}{|c|}{\textbf{4338}}                                                 & \multicolumn{1}{c|}{{\color[HTML]{333333} 172.2 (1.3)}}                                                  & \multicolumn{1}{c|}{\cellcolor[HTML]{FFCCC9}{\color[HTML]{333333} $>$ 1800}}                             & \cellcolor[HTML]{FFCCC9}{\color[HTML]{333333} $>$ 1800}                              &                       & \cellcolor[HTML]{FFFFFF}\textbf{}                                                  &                                                                                   &                                                                                   &                                                                                    &  \\ \cline{1-4}
\multicolumn{1}{|c|}{\textbf{10955}}                                                & \multicolumn{1}{c|}{\cellcolor[HTML]{FFCCC9}{\color[HTML]{333333} $>$ 1800}}                             & \multicolumn{1}{c|}{\cellcolor[HTML]{FFCCC9}{\color[HTML]{333333} $>$ 1800}}                             & \cellcolor[HTML]{FFCCC9}{\color[HTML]{333333} $>$ 1800}                              &                       & \cellcolor[HTML]{FFFFFF}\textbf{}                                                  &                                                                                   &                                                                                   &                                                                                    &  \\ \cline{1-4}
\end{tabular}}
\label{timingtable.fig}
\caption{Timing results in seconds. The fastest methods are highlighted in green and red cells indicate that the method did not finish within 30 minutes.}
\end{table}

Table 1 shows the results of our experiment across various problem sizes. We see that GBCD is much faster than Scikit-learn RuleFit (SKLRF) and Node Harvest, up to 40$\times$ faster on high dimensional problems. In addition, GBCD with warm start continuation computes the entire regularization path around 10$\times$ faster than SKLRF. 

We think that a main reason behind GBCD outperforming the SKLRF LASSO solver is that we exploit the block-structure of the problem. The leaf nodes (coordinates) in a tree ensemble are naturally grouped into trees (blocks). As tree depth increases, the number of coordinates explodes exponentially, but the number of blocks remains the same. GBCD updates blocks instead of coordinates and leverages greedy block selection heuristics, while SKLRF relies on cyclic coordinate descent. As a result, GBCD computes solutions much faster than SKLRF. Computation times of GBCD on problem \eqref{main_prob} with the MCP and fusion penalties are shown in the supplement (suppl. E).


\subsubsection{Large Problems}

As an aside, we also compare the computation time of GBCD against SKLRF for much larger problems. We modify the experimental setup in the section above to extract rules from depth 20 random forests. The corresponding optimization problems contain hundreds of thousands to millions of decision variables. Table 2 shows the results of this timing experiment; the computation time of GBCD is still much faster than the computation time of SKLRF. For the largest problem instance (10955 row dataset, >1 million decision rules), SKLRF fails to reach a solution after a day of computation. GBCD on the other hand reaches a good solution in hours. Our specialized GBCD algorithm allows \textsc{Fire} to extract decision rules from problem instances beyond the capabilities of existing off-the-shelf solvers.
\begin{table}[h]
\scalebox{0.75} {
\begin{tabular}{|ccc|}
\hline
\multicolumn{3}{|c|}{\textbf{Rule Extraction Depth 20 Ensemble}}                                                                                                       \\ \hline
\rowcolor[HTML]{C0C0C0} 
\multicolumn{1}{|c|}{\cellcolor[HTML]{C0C0C0}\textbf{Rows}} & \multicolumn{1}{c|}{\cellcolor[HTML]{C0C0C0}\textbf{GBCD}} & \textbf{Scikit-learn RuleFit}               \\ \hline
\multicolumn{1}{|c|}{\textbf{1316}}                         & \multicolumn{1}{c|}{\cellcolor[HTML]{9AFF99}8.1 mins}      & \cellcolor[HTML]{FFCCC9}56.7 mins           \\ \hline
\multicolumn{1}{|c|}{\textbf{4338}}                         & \multicolumn{1}{c|}{\cellcolor[HTML]{9AFF99}28.4 mins}     & \cellcolor[HTML]{FFCCC9}14 hrs              \\ \hline
\multicolumn{1}{|c|}{\textbf{10955}}                        & \multicolumn{1}{c|}{\cellcolor[HTML]{9AFF99}2 hrs}         & \cellcolor[HTML]{FFCCC9}\textgreater 24 hrs \\ \hline
\end{tabular}
}
\caption{Computation time of GBCD v. SKLRF for extracting rules from depth 20 tree ensembles.}
\end{table}

\section{Performance Experiments}

In this section, we compare the performance of \textsc{Fire} against competing state-of-the-art algorithms for building rule ensembles. We evaluate \textsc{Fire} against RuleFit in greater detail to better understand the effects of the MCP and fused LASSO penalties on rule extraction.

\subsection{\textsc{Fire} v. Competing Methods}

To evaluate the performance of \textsc{Fire}, we design an experiment to recreate how rule ensembles are used in practice. Rule ensembles are typically used in situations where model interpretability and transparency are important. In these situations, for a rule ensemble to be useful, the set of extracted rules must be \emph{human readable}. As such, we restrict our extracted rule set to contain less than 15 rules, with a maximum interaction depth of 3. We use \textsc{Fire} and our competing methods to extract rule ensembles under these parameters and compare the test performances of the selected models.

We repeat this procedure on 25 datasets from the OpenML repository \cite{OpenML2013} using 5-fold cross validations; the full list of datasets with metadata can be found in the supplement. First, fit a random forest of 500 depth 3 trees. Initialize \textsc{Fire} with the MCP penalty and fusion penalty. Since we are interested in selecting very sparse subsets of decision rules, we set concavity parameter $\gamma = 1.1$ close to $1^{+}$ as discussed in \S\ref{param_selection.section}. We are only interested in the performance of the selected sparse ensemble for this experiment, so we set fusion parameter $\lambda_f = 0.1$ to a low constant value. We use GBCD with warm start continuation to compute the entire regularization path for $\lambda_s$ under these \textsc{Fire} configurations. Select the value of $\lambda_s$ that produces the best model, evaluated on a validation set, of less than 15 decision rules. Record the test performance of the selected model.

We compare the performance of the model above against the following competing algorithms: RuleFit, GLRM with and without debiasing, and SIRUS. For RuleFit, we extract decision rules from the same tree ensemble as \textsc{Fire}. We tune the LASSO parameter for this algorithm on the validation set to select the best rule ensemble with less than 15 rules. SIRUS builds stabilized tree ensembles and GLRM builds decision rules using column generation. For these state-of-the-art competing algorithms, we again tune their sparsity hyperparameters on a validation set to find the best performing-rule ensemble with less than 15 rules. We record the test performances of the competing methods and compare them against \textsc{Fire}.

\begin{figure}[h]
    \centering
    \includegraphics[width = 0.4\textwidth]{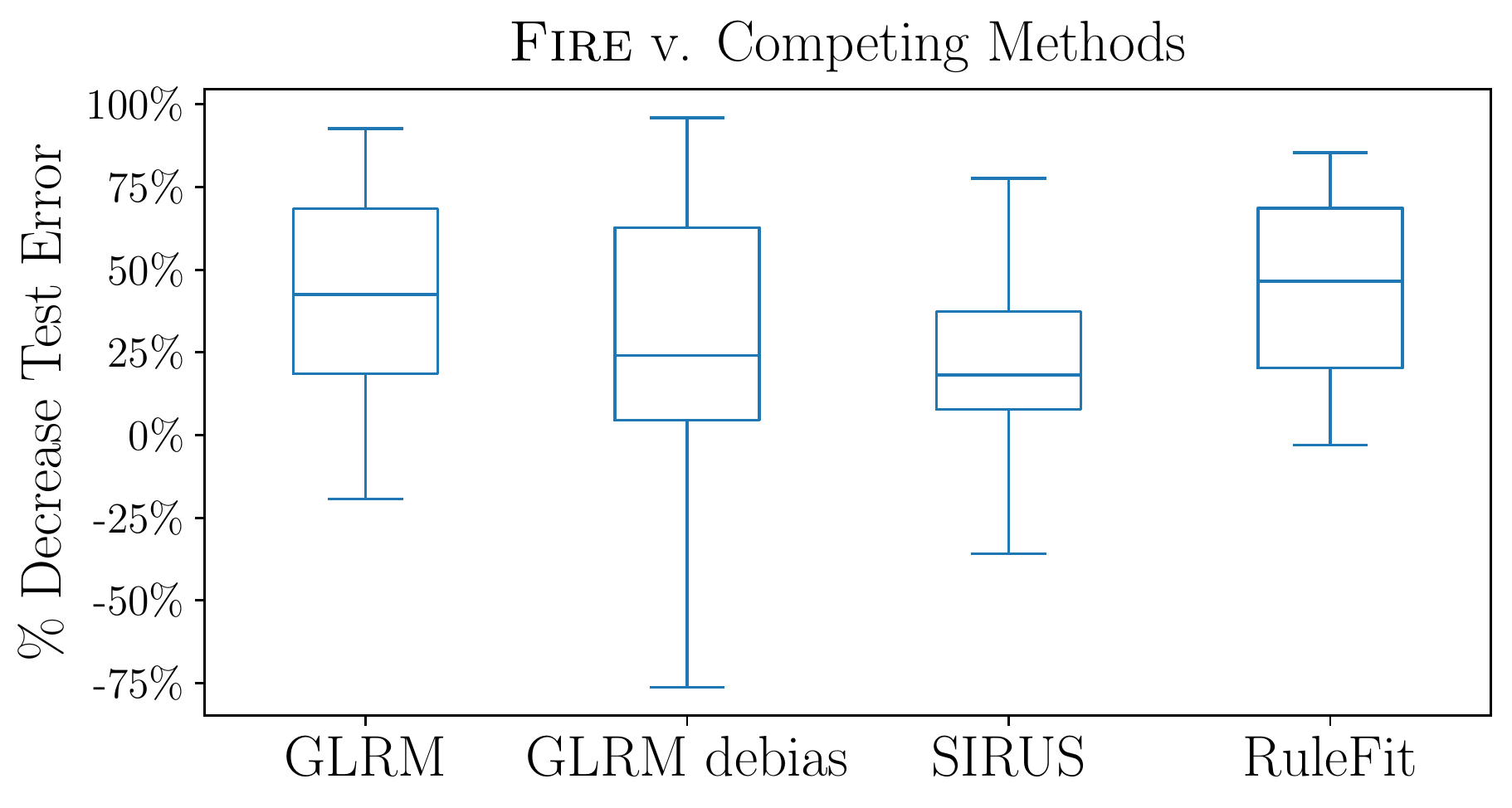}
    \caption{\textsc{Fire} outperforms SOTA competing algorithms at selecting sparse human readable rule ensembles.}
    \label{sota_competing_algorithm.fig}
\end{figure}

Figure \ref{sota_competing_algorithm.fig} presents the results of our experiment. The vertical axes show the percent decrease in test error between \textsc{Fire} and our competing methods, large positive values indicate that \textsc{Fire} performs better than our competing algorithms. The distributions of each boxplot in the figure are obtained across all datasets and folds in our experiment. We observe that the IQRs of all of the boxplots are positive. This indicates that \textsc{Fire} consistently performs better than our competing algorithms with median percent decreases in test error of \textbf{42\%} compared to GLRM without debiasing, \textbf{24\%} compared to GLRM with debiasing, \textbf{18\%} compared to SIRUS, and \textbf{46\%} compared to RuleFit. These results strongly suggest that \textsc{Fire} is a competitive algorithm for extracting sparse decision rule sets compared to state-of-the-art methods. 

One interesting thing to note is that the optional debiasing step in GLRM, where the rules are re-weighted after generation, greatly improves the performance of the algorithm. The improvement of GLRM over RuleFit found in Wei et al. (2019) \cite{wei2019generalized} may be partially due to this step since the LASSO selection in RuleFit introduces bias. We are encouraged to observe that \textsc{Fire} can outperform GLRM even with debiasing and SIRUS, two recently developed state-of-the-art algorithms for building rule ensembles.

\subsection{Further Analysis of \textsc{Fire} v. RuleFit}
Our goal here is to understand how the MCP and fusion penalties in \textsc{Fire} affect extracting decision rules from tree ensembles. To highlight the effect of our new penalties, we design this experiment to compare \textsc{Fire} with MCP and fusion against RuleFit, which extracts rules using only the LASSO penalty, across various problem sizes.

On the same datasets and folds mentioned in the section above, we fit random forests of 500 trees of depths 3,5, and 7. We initialize two versions of \textsc{Fire}. For the first version (MCP only), we set $\gamma = 1.1$ and $\lambda_f = 0$ and use GBCD with warm start continuation to compute the entire regularization path for $\lambda_s$. This version of \textsc{Fire} only uses the MCP penalty, and since $\gamma$ is close to $1^{+}$ the penalty performs aggressive selection. For the second version (MCP w/ fusion), we set $\gamma = 1.1$ and set the fusion hyperparameter $\lambda_f = 0.5\lambda_s$. Again we use warm start continuation to compute the entire regularization path for $\lambda_s$. This version of \textsc{Fire} applies a small fusion penalty which works in conjunction with the aggressive selection encouraged by the $\gamma = 1.1$ MCP penalty. For both versions of \textsc{Fire}, we record the test performance of the extracted ensemble across various sparsity levels. We compare the test performances of \textsc{Fire} against RuleFit, computed along the regularization path for the sparsity parameter. 

\begin{figure}[h]
    \centering
    \includegraphics[width = 0.45\textwidth]{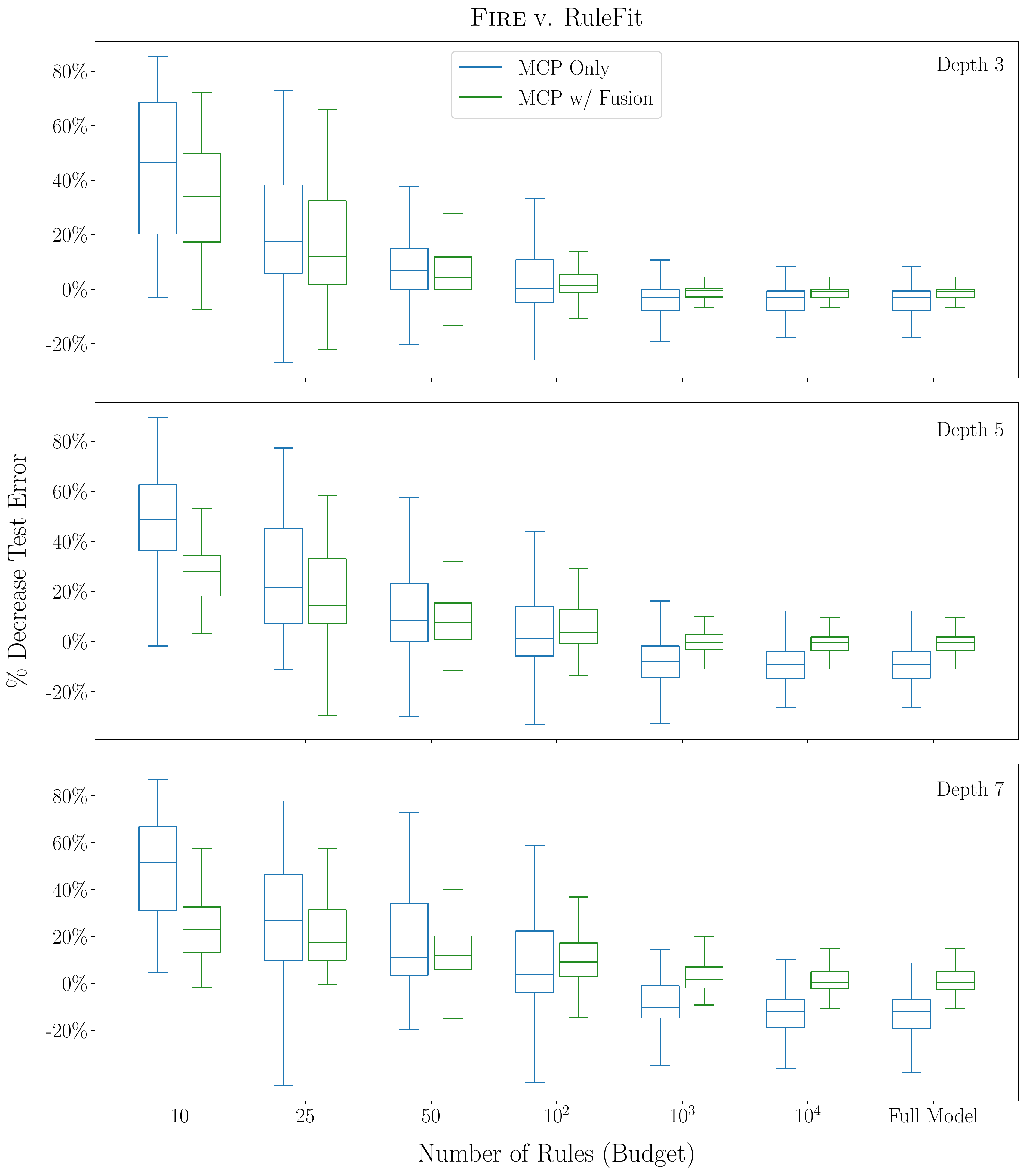}
    \caption{\textsc{Fire} v. RuleFit across various problem sizes. The MCP penalty in \textsc{Fire} performs better at selecting sparse sets of rules and the fusion penalty helps prevent overfitting.}
    \label{fire_v_rf.fig}
\end{figure}

Figure \ref{fire_v_rf.fig} shows the results of this experiment. The plots compare the best model selected by \textsc{Fire} against the best model selected by RuleFit given a budget or rules, shown on the horizontal axis. The vertical axes show the percent decrease in test error between \textsc{Fire} and RuleFit, values above $0\%$ indicate that \textsc{Fire} performs better than RuleFit. The distributions for each boxplot are again obtained across all folds and datasets in the experiment. We observe that \textsc{Fire} with the MCP penalties perform substantially better than RuleFit at selecting sparse models (LHS of figure \ref{fire_v_rf.fig}). This is expected due to the behavior of the MCP penalty compared to the LASSO penalty when $\gamma \xrightarrow[]{} 1^{+}$. We also note that \textsc{Fire} with the MCP penalty only performs slightly better than \textsc{Fire} with both the MCP and fusion penalty in this regime. This is likely due to the fact that the additional regularization imparted by the fusion penalty causes the model to underfit when performing very sparse selection. Consequently, we suggest keeping $\lambda_f$ small when extracting sparse rule sets.

When the sparsity regularization penalty is reduced (i.e., the rule budget is increased) we observe that \textsc{Fire} with the MCP penalty only begins to overfit. This effect is especially pronounced when extracting decision rules from the depth 7 tree ensemble (bottom panel of figure \ref{fire_v_rf.fig}), due to the inherent complexity of the deeper rules. \textsc{Fire} with both the MCP and fusion penalty avoids this issue since the fusion penalty adds additional regularization. We see in figure \ref{fire_v_rf.fig} that \textsc{Fire} with both the MCP and fusion penalty outperform RuleFit across all rule budgets; all of the green boxplots in the figure lie above 0\%. By combining the aggressive selection of the MCP penalty with the regularization added by fusion, \textsc{Fire} outperforms RuleFit at extracting rule ensembles across all model sparsities.

\section{Interpretability Case Study}
\label{interpretability_case.section}

\begin{figure}[h]
    \centering
    \includegraphics[width = 0.48\textwidth]{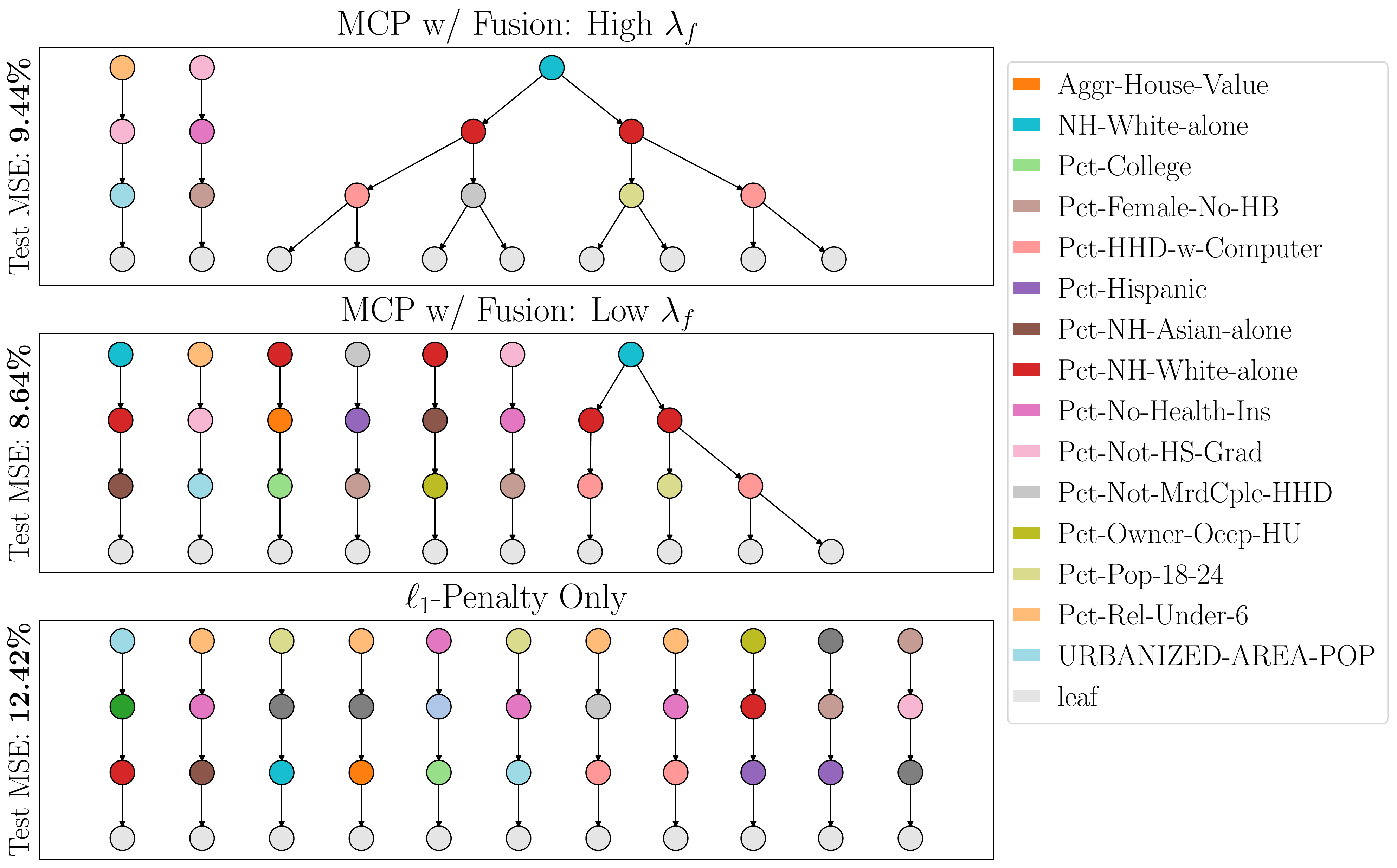}
    \caption{US Census case study:
    \textsc{Fire} extracts more interpretable, better performing ensembles compared to RuleFit.}
    \label{census_example.fig}
\end{figure}

We conclude with a case study to showcase the improved interpretability of $\textsc{Fire}$ in a real-world example. We follow the work by Ibrahim et al. (2021) \cite{ibrahim2021predicting} and use the Census Planning Database to predict census tract response rates. The US Census Bureau wants to understand what features influence response rate so that low-response tracts can be targeted; previous efforts have found that tree ensembles perform well but are difficult to interpret \citep{erdman2017low}. We use \textsc{Fire} to extract an interpretable set of decision rules.

We first build a random forest of 500 depth 3 trees. The full model achieves a test MSE of \textbf{8.64\%}. We use \textsc{Fire} with the MCP and fusion penalties to extract around 10 decision rules with $\gamma = 1.1$ and $\lambda_f = 0.5 \lambda_s$ (low) or $\lambda_f = 2 \lambda_s$ (high). The 10 rules selected by \textsc{Fire} with low $\gamma_f$ perform the same as the full ensemble (\textbf{8.64\%} MSE) and the 10 rules selected when $\lambda_f$ is high perform slightly worse (\textbf{9.44\%} MSE). In contrast, when we select 11 rules with RuleFit, the model performs much worse (\textbf{12.42\%} MSE).

The rule ensembles extracted by \textsc{Fire} are substantially more interpretable than the RuleFit ensembles. The bottom plot in figure \ref{census_example.fig} shows the 11 rules selected by RuleFit. These rules are selected across different trees and are not grouped in any meaningful manner; it is difficult to interpret this model from the figure alone. In comparison, the middle plot contains the 10 rule ensemble selected by \textsc{Fire} with low $\lambda_f$. This rule ensemble contains a partial decision tree which reveals that the feature \textit{NH-White-alone} is important since 4 rules share the same split on this feature. Increasing $\lambda_f$ yields the most interpretable rule ensemble, shown in the top plot, which consists of a single decision tree and 2 additional rules.

As an attempt to quantify interpretability, we can count the number of antecedents, colored nodes in figure \ref{census_example.fig}, that a user must analyze to interpret a rule ensemble. The RuleFit ensemble contains \textbf{33} antecedents while the \textsc{Fire} high $\lambda_f$ ensemble contains just \textbf{13}. 


\section{Conclusion}

\textsc{Fire} is a novel optimization-based framework to extract decision rules from tree ensembles. The framework selects sparse representative subsets of rules from an ensemble and allows for the flexibility to encourage rule fusion during the selection procedure. This improves the interpretability and compression of the extracted model since many of the selected rules share common antecedents. \textsc{Fire} uses a non-convex MCP penalty to aggressively select rules in the presence of correlations and a fused LASSO penalty to encourage rule fusion. To solve the large non-convex optimization problems in \textsc{Fire}, we develop a specialized GBCD solver that computes high-quality solutions efficiently. Our solver exploits the blocking structure of the problem and leverages greedy block selection heuristics to reduce computation time. As a result, our solver scales well and allows for computation beyond the capabilities of off-the-shelf methods. Our experiments show that \textsc{Fire} performs better than state-of-the-art algorithms at building human readable rule sets and that \textsc{Fire}
outperforms RuleFit at extracting rule ensembles across all sparsity levels. Altogether, these features and finding make \textsc{Fire} a fast and effective framework for extracting interpretable rule ensembles.

\textbf{ACKNOWLEDGMENTS} This research is funded in part by a grant from the Office of Naval Research (ONR-N00014-21-1-2841).

\bibliographystyle{plainnat}
\balance
\bibliography{ref2}

\begin{thebibliography}{30}
\providecommand{\natexlab}[1]{#1}
\providecommand{\url}[1]{\texttt{#1}}
\expandafter\ifx\csname urlstyle\endcsname\relax
  \providecommand{\doi}[1]{doi: #1}\else
  \providecommand{\doi}{doi: \begingroup \urlstyle{rm}\Url}\fi

\bibitem[Beck and Teboulle(2009)]{beck2009fast}
Amir Beck and Marc Teboulle.
\newblock A fast iterative shrinkage-thresholding algorithm for linear inverse
  problems.
\newblock \emph{SIAM journal on imaging sciences}, 2\penalty0 (1):\penalty0
  183--202, 2009.

\bibitem[B{\'e}nard et~al.(2021)B{\'e}nard, Biau, Da~Veiga, and
  Scornet]{benard2021sirus}
Cl{\'e}ment B{\'e}nard, G{\'e}rard Biau, S{\'e}bastien Da~Veiga, and Erwan
  Scornet.
\newblock Sirus: Stable and interpretable rule set for classification.
\newblock \emph{Electronic Journal of Statistics}, 15\penalty0 (1):\penalty0
  427--505, 2021.

\bibitem[Breiman(1996)]{breiman1996bagging}
Leo Breiman.
\newblock Bagging predictors.
\newblock \emph{Machine learning}, 24\penalty0 (2):\penalty0 123--140, 1996.

\bibitem[Diamond and Boyd(2016)]{diamond2016cvxpy}
Steven Diamond and Stephen Boyd.
\newblock {CVXPY}: {A} {P}ython-embedded modeling language for convex
  optimization.
\newblock \emph{Journal of Machine Learning Research}, 17\penalty0
  (83):\penalty0 1--5, 2016.

\bibitem[Erdman and Bates(2017)]{erdman2017low}
Chandra Erdman and Nancy Bates.
\newblock The low response score (lrs) a metric to locate, predict, and manage
  hard-to-survey populations.
\newblock \emph{Public Opinion Quarterly}, 81\penalty0 (1):\penalty0 144--156,
  2017.

\bibitem[Fan and Li(2001)]{fan2001variable}
Jianqing Fan and Runze Li.
\newblock Variable selection via nonconcave penalized likelihood and its oracle
  properties.
\newblock \emph{Journal of the American statistical Association}, 96\penalty0
  (456):\penalty0 1348--1360, 2001.

\bibitem[Fang et~al.(2020)Fang, Fan, Sun, and Friedlander]{fang2020greed}
Huang Fang, Zhenan Fan, Yifan Sun, and Michael Friedlander.
\newblock Greed meets sparsity: Understanding and improving greedy coordinate
  descent for sparse optimization.
\newblock In \emph{International Conference on Artificial Intelligence and
  Statistics}, pages 434--444. PMLR, 2020.

\bibitem[Friedman et~al.(2007)Friedman, Hastie, H{\"o}fling, and
  Tibshirani]{friedman2007pathwise}
Jerome Friedman, Trevor Hastie, Holger H{\"o}fling, and Robert Tibshirani.
\newblock Pathwise coordinate optimization.
\newblock \emph{The annals of applied statistics}, 1\penalty0 (2):\penalty0
  302--332, 2007.

\bibitem[Friedman et~al.(2010)Friedman, Hastie, and
  Tibshirani]{friedman2010regularization}
Jerome Friedman, Trevor Hastie, and Rob Tibshirani.
\newblock Regularization paths for generalized linear models via coordinate
  descent.
\newblock \emph{Journal of statistical software}, 33\penalty0 (1):\penalty0 1,
  2010.

\bibitem[Friedman(2001)]{friedman2001greedy}
Jerome~H Friedman.
\newblock Greedy function approximation: a gradient boosting machine.
\newblock \emph{Annals of statistics}, pages 1189--1232, 2001.

\bibitem[Friedman and Popescu(2008)]{friedman2008predictive}
Jerome~H Friedman and Bogdan~E Popescu.
\newblock Predictive learning via rule ensembles.
\newblock \emph{The annals of applied statistics}, pages 916--954, 2008.

\bibitem[Hazimeh et~al.(2022)Hazimeh, Mazumder, and Nonet]{hazimeh2022l0learn}
Hussein Hazimeh, Rahul Mazumder, and Tim Nonet.
\newblock L0learn: A scalable package for sparse learning using l0
  regularization.
\newblock \emph{arXiv preprint arXiv:2202.04820}, 2022.

\bibitem[Hebiri and Lederer(2012)]{hebiri2012correlations}
Mohamed Hebiri and Johannes Lederer.
\newblock How correlations influence lasso prediction.
\newblock \emph{IEEE Transactions on Information Theory}, 59\penalty0
  (3):\penalty0 1846--1854, 2012.

\bibitem[Hoefling(2010)]{hoefling2010path}
Holger Hoefling.
\newblock A path algorithm for the fused lasso signal approximator.
\newblock \emph{Journal of Computational and Graphical Statistics}, 19\penalty0
  (4):\penalty0 984--1006, 2010.

\bibitem[Ibrahim et~al.(2021)Ibrahim, Mazumder, Radchenko, and
  Ben-David]{ibrahim2021predicting}
Shibal Ibrahim, Rahul Mazumder, Peter Radchenko, and Emanuel Ben-David.
\newblock Predicting census survey response rates via interpretable
  nonparametric additive models with structured interactions.
\newblock \emph{arXiv preprint arXiv:2108.11328}, 2021.

\bibitem[Johnson(2013)]{johnson2013dynamic}
Nicholas~A Johnson.
\newblock A dynamic programming algorithm for the fused lasso and l
  0-segmentation.
\newblock \emph{Journal of Computational and Graphical Statistics}, 22\penalty0
  (2):\penalty0 246--260, 2013.

\bibitem[Karimireddy et~al.(2019)Karimireddy, Koloskova, Stich, and
  Jaggi]{karimireddy2019efficient}
Sai~Praneeth Karimireddy, Anastasia Koloskova, Sebastian~U Stich, and Martin
  Jaggi.
\newblock Efficient greedy coordinate descent for composite problems.
\newblock In \emph{The 22nd International Conference on Artificial Intelligence
  and Statistics}, pages 2887--2896. PMLR, 2019.

\bibitem[Mazumder et~al.(2011)Mazumder, Friedman, and
  Hastie]{mazumder2011sparsenet}
Rahul Mazumder, Jerome~H Friedman, and Trevor Hastie.
\newblock Sparsenet: Coordinate descent with nonconvex penalties.
\newblock \emph{Journal of the American Statistical Association}, 106\penalty0
  (495):\penalty0 1125--1138, 2011.

\bibitem[Meinshausen(2010)]{meinshausen2010node}
Nicolai Meinshausen.
\newblock Node harvest.
\newblock \emph{The Annals of Applied Statistics}, pages 2049--2072, 2010.

\bibitem[Molnar(2016)]{molnar_2016}
Christopher Molnar.
\newblock Rulefit: Python implementation of the rulefit algorithm, Jun 2016.
\newblock URL \url{https://github.com/christophM/rulefit}.

\bibitem[Nesterov et~al.(2018)]{nesterov2018lectures}
Yurii Nesterov et~al.
\newblock \emph{Lectures on convex optimization}, volume 137.
\newblock Springer, 2018.

\bibitem[Nutini et~al.(2015)Nutini, Schmidt, Laradji, Friedlander, and
  Koepke]{nutini2015coordinate}
Julie Nutini, Mark Schmidt, Issam Laradji, Michael Friedlander, and Hoyt
  Koepke.
\newblock Coordinate descent converges faster with the gauss-southwell rule
  than random selection.
\newblock In \emph{International Conference on Machine Learning}, pages
  1632--1641. PMLR, 2015.

\bibitem[Nutini et~al.(2017)Nutini, Laradji, and Schmidt]{nutini2017let}
Julie Nutini, Issam Laradji, and Mark Schmidt.
\newblock Let's make block coordinate descent go fast: Faster greedy rules,
  message-passing, active-set complexity, and superlinear convergence.
\newblock \emph{arXiv preprint arXiv:1712.08859}, 2017.

\bibitem[Pedregosa et~al.(2011)Pedregosa, Varoquaux, Gramfort, Michel, Thirion,
  Grisel, Blondel, Prettenhofer, Weiss, Dubourg, Vanderplas, Passos,
  Cournapeau, Brucher, Perrot, and Duchesnay]{scikit-learn}
F.~Pedregosa, G.~Varoquaux, A.~Gramfort, V.~Michel, B.~Thirion, O.~Grisel,
  M.~Blondel, P.~Prettenhofer, R.~Weiss, V.~Dubourg, J.~Vanderplas, A.~Passos,
  D.~Cournapeau, M.~Brucher, M.~Perrot, and E.~Duchesnay.
\newblock Scikit-learn: Machine learning in {P}ython.
\newblock \emph{Journal of Machine Learning Research}, 12:\penalty0 2825--2830,
  2011.

\bibitem[Sun et~al.(2020)Sun, Chain, Kaski, and
  Shawe-Taylor]{sun2020correlated}
Yuxin Sun, Benny Chain, Samuel Kaski, and John Shawe-Taylor.
\newblock Correlated feature selection with extended exclusive group lasso.
\newblock \emph{arXiv preprint arXiv:2002.12460}, 2020.

\bibitem[Tibshirani et~al.(2005)Tibshirani, Saunders, Rosset, Zhu, and
  Knight]{tibshirani2005sparsity}
Robert Tibshirani, Michael Saunders, Saharon Rosset, Ji~Zhu, and Keith Knight.
\newblock Sparsity and smoothness via the fused lasso.
\newblock \emph{Journal of the Royal Statistical Society: Series B (Statistical
  Methodology)}, 67\penalty0 (1):\penalty0 91--108, 2005.

\bibitem[Tibshirani and Taylor(2011)]{tibshirani2011solution}
Ryan~J Tibshirani and Jonathan Taylor.
\newblock The solution path of the generalized lasso.
\newblock \emph{The annals of statistics}, 39\penalty0 (3):\penalty0
  1335--1371, 2011.

\bibitem[Vanschoren et~al.(2013)Vanschoren, van Rijn, Bischl, and
  Torgo]{OpenML2013}
Joaquin Vanschoren, Jan~N. van Rijn, Bernd Bischl, and Luis Torgo.
\newblock Openml: networked science in machine learning.
\newblock \emph{SIGKDD Explorations}, 15\penalty0 (2):\penalty0 49--60, 2013.
\newblock \doi{10.1145/2641190.2641198}.
\newblock URL \url{http://doi.acm.org/10.1145/2641190.264119}.

\bibitem[Wei et~al.(2019)Wei, Dash, Gao, and Gunluk]{wei2019generalized}
Dennis Wei, Sanjeeb Dash, Tian Gao, and Oktay Gunluk.
\newblock Generalized linear rule models.
\newblock In \emph{International Conference on Machine Learning}, pages
  6687--6696. PMLR, 2019.

\bibitem[Zhang(2010)]{zhang2010nearly}
Cun-Hui Zhang.
\newblock Nearly unbiased variable selection under minimax concave penalty.
\newblock \emph{The Annals of statistics}, 38\penalty0 (2):\penalty0 894--942,
  2010.

\end{thebibliography}

\end{document}